\documentclass[11pt]{article}

\usepackage[margin=1in]{geometry}

\usepackage[T1]{fontenc}
\usepackage{times}

\usepackage{amsmath}
\usepackage{amssymb}

\usepackage{booktabs}
\usepackage{tabularx}
\usepackage{array}
\usepackage{multirow}

\usepackage{graphicx}
\usepackage{subcaption}
\usepackage{float}
\usepackage{placeins}

\usepackage{enumitem}
\usepackage{needspace}

\usepackage[numbers,sort&compress]{natbib}

\usepackage[hyphens]{url}
\usepackage[colorlinks=true,citecolor=blue,linkcolor=blue,urlcolor=blue]{hyperref}
\title{\textbf{BRIDGE: Bottleneck-Aware Regulator-Set Inference and Diagnosis for Cooperative Gene Regulatory Recovery}}

\author{
Maryam Rahimimovassagh\\
University of Central Florida\\
\texttt{maryam.rahimimovassagh@ucf.edu}
\and
Clayton Barham\\
University of Central Florida\\
\texttt{cl964111@ucf.edu}
\and
Ivan Garibay\\
University of Central Florida\\
\texttt{igaribay@ucf.edu}
\and
Waldemar Karwowski\\
University of Central Florida\\
\texttt{wkar@ucf.edu}
\and
Niloofar Yousefi\\
University of Central Florida\\
\texttt{niloofar.yousefi@ucf.edu}
}

\date{}

\begin{document}

\maketitle

\begin{abstract}
Cooperative gene regulation often depends on groups of regulators acting
jointly, but most gene regulatory network (GRN) inference methods output
pairwise regulator--target rankings. We introduce Bottleneck-Aware
Regulator-Set Inference and Diagnosis (BRIDGE), a framework for complete
regulator-set recovery, and Targeted Recovery Attribution for Cooperative
Evaluation (TRACE), a diagnostic suite that attributes failures to retrieval,
set-level scoring, decoding, and evaluation bottlenecks.

TRACE includes a leak-free mechanism-mismatch cooperativity stress test in
which cooperative targets are generated by random nonlinear mechanisms rather
than product interactions. This design avoids feature--mechanism circularity:
Residual higher-order set scoring (Residual HOS2) operates on raw expression
vectors without handcrafted product-correlation features. In this stress test,
Residual HOS2 consistently improves graded recovery over a decomposable
pairwise set scorer (PairS2) across all matched seed--cooperativity settings.
Averaged over all \(30\) seed--cooperativity settings, Jaccard improves from
\(0.382\) to \(0.460\), recall from \(0.522\) to \(0.597\), and exact recovery
improves on average from \(0.053\) to \(0.113\). However, exact recovery remains
low, showing that complete regulator-set recovery remains a strict bottleneck.

On SERGIO DS3, oracle retrieval and formal TRACE attribution show that candidate
coverage is necessary but not sufficient: set-level misranking remains the
dominant source of exact-recovery failure. BRIDGE further shows that interventions are
validation-dependent, with PairS2 proposal followed by Residual HOS2 reranking
reducing the number of HOS2-scored candidate sets by approximately
\(94\)--\(97\%\) while largely preserving exact-recovery behavior. Together,
these results separate edge ranking, candidate retrieval, set-level scoring,
and exact cooperative regulator-set recovery as distinct objectives.
\end{abstract}
\begin{figure*}[t]
    \centering
    \includegraphics[
        width=\textwidth,
        trim={10pt 5pt 45pt 10pt},
        clip
    ]{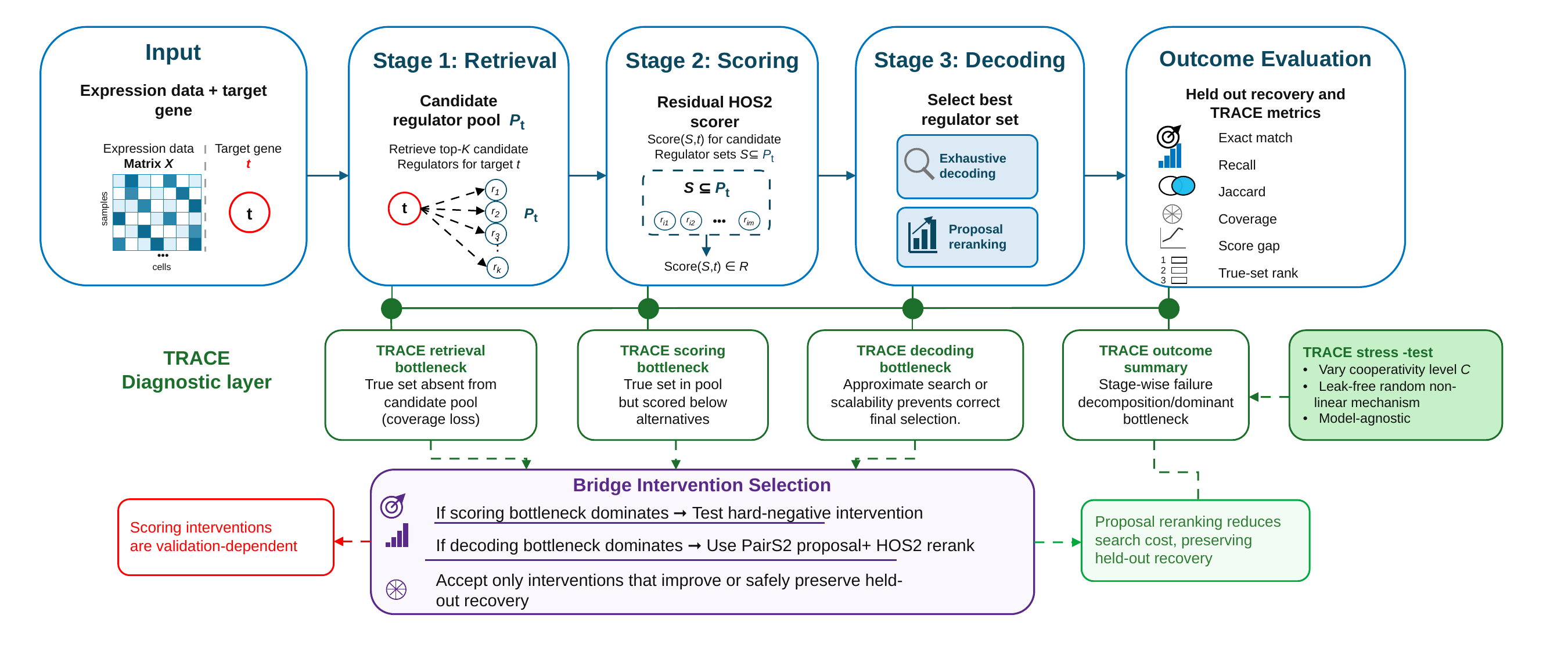}
    \caption{
        BRIDGE/TRACE bottleneck-aware recovery pipeline. BRIDGE decomposes
        cooperative regulator-set recovery into retrieval, scoring, decoding,
        and outcome-evaluation stages. TRACE attributes failures to
        stage-specific bottlenecks, and BRIDGE uses these diagnoses to accept
        only interventions that improve or safely preserve held-out recovery.
    }
    \label{fig:bridge-trace-framework}
\end{figure*}

\section{Introduction}

Gene regulatory network (GRN) inference seeks to recover regulatory
relationships governing gene expression from high-dimensional molecular data.
GRNs are central to developmental control, cellular identity, and transcriptional
regulation \cite{davidson2005grn,lambert2018tfs}. Most GRN inference methods
formulate this task as ranking pairwise regulator--target associations
\(r_i \rightarrow t\), using mutual-information, tree-based, boosting,
single-cell workflow, dynamical, regression, deep-learning, perturbation, and
contrastive approaches
\cite{Margolin2006ARACNE,HuynhThu2010GENIE3,Moerman2019GRNBoost2,Aibar2017SCENIC,
Chan2017PIDC,Matsumoto2017SCODE,Gibbs2022Inferelator,Shu2021DeepSEM,
Kamimoto2023CellOracle,Roohani2024GEARS,Zhao2025BEACON}. In parallel,
single-cell foundation models have expanded transcriptomic representation
learning through large-scale pretraining and transfer
\cite{Yang2022scBERT,Theodoris2023Geneformer,Cui2024scGPT,Hao2024scFoundation}.
However, edge-ranking performance, representation quality, and cooperative
regulator-set recovery are distinct objectives, and benchmarking studies show
that single-cell GRN inference remains strongly dependent on modeling
assumptions, data modality, and evaluation protocol
\cite{Marbach2012DREAM5,Pratapa2020Benchmarking}.

This creates a mismatch with the biological structure of cooperative regulation.
Transcription factors rarely act alone. Many targets are controlled by groups of
regulators acting jointly:
\[
S_t = \{r_1, r_2, \ldots, r_k\}.
\]
Enhancer-mediated regulation often depends on combinations of transcription
factors, cofactors, chromatin context, motif grammar, binding-site orientation,
and cooperative binding
\cite{Spitz2012Transcription,panne2007enhanceosome,mirny2010nucleosome,
reiter2017combinatorial,rao2021cooperative,georgakopoulos2023tfbs}.
More broadly, higher-order interactions can encode group effects that are not
captured by pairwise links alone
\cite{benson2016higher,Battiston2020Networks}. Recent hypergraph-based GRN
methods further reflect growing interest in representing regulatory structure
beyond pairwise edges
\cite{su2025hypergraphgrn}.

The key difficulty is that strong edge-level performance does not necessarily
imply recovery of the underlying regulatory mechanism. A method may rank several
true regulators highly while still failing to recover the exact functional team.
Recovering most, but not all, members of a regulator set can yield an incomplete
or misleading mechanism. This failure mode is amplified by hub or transitive
confounders, where genes with strong marginal association dominate pairwise
rankings even when the target is governed by a specific interacting regulator
set.

This paper studies cooperative regulator-set recovery as a set-structured
scientific discovery problem: given a target gene \(t\), recover the complete
regulator set \(S_t\), rather than only a ranked list of regulator--target
edges. This objective exposes distinct failure modes: the true regulators may
never enter the candidate pool, the correct set may be scored below a near-miss
alternative, the decoder may miss the relevant subset space, or edge-level
evaluation may hide set-level failure. \emph{This raises the central question:
do current methods fail because they cannot learn cooperative biological
structure, or because retrieval, scoring, decoding, and evaluation bottlenecks
obscure recoverable information?}

This work builds on recent residual set modeling for combinatorial gene
regulation~\cite{Rahimimovassagh2026Beyond}, but shifts the focus from
proposing a higher-order set scorer to diagnosing why cooperative regulator-set
recovery fails across retrieval, scoring, decoding, and evaluation stages. We
introduce BRIDGE, a Bottleneck-aware Regulator-Set Inference and Diagnosis
framework for cooperative gene regulatory recovery. BRIDGE decomposes exact
regulator-set recovery into candidate retrieval, set-level scoring,
combinatorial decoding, and outcome evaluation. Its diagnostic component,
TRACE, attributes recovery failures to stage-specific bottlenecks using
measurable quantities such as coverage, conditional exact recovery, score gap,
true-set rank, runtime, and approximate-decoder agreement.

Within TRACE, we include a leak-free mechanism-mismatch cooperativity stress
test and a SERGIO DS3 bottleneck analysis. The stress test replaces
product-based mechanisms with target-specific random nonlinear mechanisms,
avoiding feature--mechanism circularity and isolating set-level scoring through
oracle-covered candidate pools. On SERGIO DS3 \cite{Dibaeinia2020SERGIO},
TRACE diagnoses how coverage, scoring, decoding cost, and exact recovery
interact. BRIDGE uses these diagnoses to compare models and guide interventions,
following the need for careful single-cell GRN benchmarking
\cite{Pratapa2020Benchmarking}.

We also define TRACE-lite for external edge-ranking baselines that do not expose
explicit retrieval, set-scoring, and decoding modules. TRACE-lite converts
ranked regulator--target outputs into fixed top-\(M\) candidate pools and
top-\(R\) predicted regulator sets, allowing methods such as correlation,
mutual information, GRNBoost2, and GENIE3 to be evaluated under the same
cooperative set-recovery protocol. This distinction is important because
edge-ranking performance, perturbation prediction, foundation-model
representation quality, and complete cooperative set recovery answer different
questions.

BRIDGE/TRACE is model-agnostic: it does not require a particular GRN architecture, feature representation, or scoring model. It provides a common protocol for testing whether a method retrieves, scores, and selects the complete regulator set for a target gene. TRACE inspects modular retrieval, scoring, and decoding stages directly, while TRACE-lite converts external edge rankings into matched top-\(M\) candidate pools and top-\(R\) predicted sets. This separates edge-ranking success, candidate coverage, set-level discrimination, and exact cooperative mechanism recovery.

\subsection{Contributions}

This work makes the following contributions:

\begin{enumerate}[leftmargin=*]
    \item We formulate cooperative gene regulatory network (GRN) inference as
    complete regulator-set recovery rather than edge ranking.

    \item We introduce BRIDGE, a stage-wise framework that separates retrieval,
    set-level scoring, decoding, and evaluation.

    \item We introduce TRACE, a diagnostic protocol for attributing failures
    using coverage, conditional exact recovery, score gaps, true-set ranks, and
    decoder comparisons.

    \item We use Residual higher-order set scoring (Residual HOS2) as a
    set-scoring probe to test whether non-additive set modeling improves
    recovery beyond decomposable pairwise set scoring (PairS2).

    \item We show across controlled, SERGIO DS3, and real-label TRACE-lite
    experiments that regulator-level recovery and complete regulator-set
    recovery are distinct objectives.
\end{enumerate}
\begin{table*}[t]
\centering
\footnotesize
\setlength{\tabcolsep}{4pt}
\renewcommand{\arraystretch}{1.12}
\caption{Terminology used throughout the paper.}
\label{tab:terminology}
\begin{tabular}{@{}ll@{}}
\toprule
Term & Meaning \\
\midrule
GRN & Gene regulatory network. \\
BRIDGE & Bottleneck-Aware Regulator-Set Inference and Diagnosis framework. \\
TRACE & Diagnostic suite for retrieval, scoring, decoding, and evaluation bottlenecks. \\
TRACE-lite & Output-level TRACE protocol for external edge-ranking baselines. \\
Stage1TopR & Retrieval-only baseline that predicts the top-\(R\) regulators. \\
PairS2 & Decomposable pairwise set scorer that sums regulator--target scores. \\
Residual HOS2 & Residual higher-order set scorer with PairS2 backbone plus learned set correction. \\
Coverage@\(M\) & Whether the complete true regulator set appears in the top-\(M\) candidate pool. \\
Conditional Exact & Exact recovery restricted to targets whose true set is present in the candidate pool. \\
Score Gap & Difference between the true-set score and the highest-scoring incorrect-set score. \\
True-Set Rank & Rank of the true regulator set among evaluated candidate sets. \\
\bottomrule
\end{tabular}
\end{table*}

\section{Problem Setup and Operational View}

Given a target gene \(t\), cooperative regulator-set recovery aims to recover
the complete regulator set
\[
S_t=\{r_1,\ldots,r_R\},
\]
rather than only ranking individual regulator--target edges. This differs from
standard gene regulatory network (GRN) inference, where methods typically
output pairwise regulator--target scores.

A useful way to distinguish pairwise and cooperative regulation is through
additive versus non-additive set effects. In an additive regulatory model, the
target response can be approximated as a sum of independent regulator effects:
\[
f_t(S_t) \approx \sum_{r_i\in S_t} f_t(r_i).
\]
In a cooperative regulatory model, the regulator set contributes an additional
higher-order term:
\[
f_t(S_t)
=
\sum_{r_i\in S_t} f_t(r_i)
+
g_t(S_t).
\]
Here, \(g_t(S_t)\) captures the non-additive contribution of the regulator set.
When \(g_t(S_t)\neq 0\), the complete regulator set carries information that
cannot be recovered from independent regulator--target effects alone. In this
paper, cooperativity therefore refers both to a biological property of
transcriptional regulation and to an operational machine-learning target: the
recovery of regulator sets whose collective activity contains information
beyond pairwise interactions.

This set-level objective exposes several distinct failure modes. The true
regulators may be absent from the candidate pool, the true set may be present
but scored below a near-miss alternative, the decoder may fail to search the
relevant subset space, or edge-level evaluation may hide set-level failure.
BRIDGE therefore treats exact recovery as a stage-wise recoverability problem
over retrieval, scoring, decoding, and evaluation.

We use the information-bottleneck perspective only as an operational framing:
a useful representation should preserve task-relevant information while
discarding irrelevant variation. Here, the task-relevant information is the
identity of \(S_t\). Rather than estimating high-dimensional mutual information
directly, TRACE uses measurable quantities such as coverage, score gap,
true-set rank, and decoder agreement to identify where recovery-relevant
information is lost, obscured, or made inaccessible.
\section{BRIDGE Framework}
\label{sec:bridge}

BRIDGE is the umbrella framework for bottleneck-aware cooperative
regulator-set recovery. It organizes complete regulator-set recovery into
candidate retrieval, set-level scoring, combinatorial decoding, and outcome
evaluation. Within BRIDGE, TRACE provides the diagnostic suite for measuring
where recovery fails, while the BRIDGE intervention policy uses these diagnoses
to compare recovery models and guide targeted interventions.

\subsection{TRACE Diagnostic Suite}

TRACE measures stage-wise recovery using directly observable quantities. For a
target gene \(t\), let \(S_t\) denote the true regulator set, \(C_t(M)\) the
top-\(M\) candidate regulator pool, and \(R=|S_t|\) the regulator-set size.

\paragraph{Retrieval bottleneck.}
Retrieval measures whether the complete true regulator set is present in the
candidate pool:
\[
\mathrm{Coverage@}M = \Pr(S_t \subseteq C_t(M)).
\]
If \(S_t\) is absent from \(C_t(M)\), no downstream scorer or decoder can
recover the complete set. We therefore define retrieval loss as
\[
L_{\mathrm{ret}} = 1-\mathrm{Coverage@}M.
\]

\paragraph{Search-space burden.}
For a candidate pool of size \(M\) and regulator-set size \(R\), the decoder
must search among
\[
\binom{M}{R}
\]
possible regulator sets. We use
\[
H_R(M)=\log \binom{M}{R}
\]
as an operational search-space entropy, measuring the log-size of the subset
space. This motivates reporting not only exact recovery, but also candidate-set
counts and runtime.

\paragraph{Scoring bottleneck.}
Set-level scoring is evaluated among targets whose true set is present in the
candidate pool. For a covered target, let
\[
S_t^{-}
=
\arg\max_{\substack{S\subseteq C_t(M),\, |S|=R\\ S\neq S_t}}
\mathrm{Score}(S,t)
\]
be the highest-scoring incorrect regulator set. The score gap is
\[
\mathrm{Gap}_t =
\mathrm{Score}(S_t,t)-\mathrm{Score}(S_t^{-},t).
\]
A positive gap means the true set is ranked above all incorrect candidate sets;
a non-positive gap indicates a scoring failure or ambiguity. At the population
level, we report scoring loss as
\[
L_{\mathrm{score}}^{\mathrm{pop}}
=
\mathrm{Coverage@}M-\mathrm{UnconditionalExact}.
\]
Under exhaustive decoding, this quantity captures covered targets for which the
true set is present but not ranked first by the scorer.

\paragraph{True-set rank.}
The score gap gives a binary view of scoring success. We also compute the
true-set rank, defined as
\[
\mathrm{Rank}_t = 1 + N_t,
\]
where \(N_t\) is the number of incorrect candidate sets with score greater than
or equal to the score of the true set. Rank \(1\) means the true regulator set is
the highest-scoring set in the evaluated pool; larger ranks indicate stronger
set-level confusion.

\paragraph{Decoding bottleneck.}
Decoding measures whether the search procedure recovers the highest-scoring set
under computational constraints. Exhaustive decoding evaluates every size-\(R\)
subset in the capped candidate pool and therefore has zero decoding loss within
that pool. Approximate decoders, such as proposal reranking or beam search, can
introduce a decoding bottleneck when they fail to recover the exhaustive-search
solution. Additional operational definitions are provided in
Appendix~\ref{app:information_theoretic_bridge}. 

\subsection{Recovery Models}

BRIDGE compares three recovery models. Stage1TopR is a retrieval-only baseline
that predicts the top-\(R\) regulators from the Stage-1 candidate list. Pairwise
set scoring (PairS2) is a decomposable Stage-2 scorer:
\[
\mathrm{Score}_{\mathrm{PairS2}}(S,t)
=
\sum_{r\in S}\phi_{\mathrm{pair}}(r,t).
\]
Thus, PairS2 decoding reduces to selecting the top-\(R\) individually scored
regulators.

Residual higher-order set scoring (Residual HOS2) extends PairS2 with a learned
non-additive set correction:
\[
\mathrm{Score}_{\mathrm{HOS2}}(S,t)
=
\sum_{r\in S}\phi_{\mathrm{pair}}(r,t)+\psi_{\mathrm{set}}(S,t).
\]
The correction \(\psi_{\mathrm{set}}\) is implemented by a target-conditioned
Set Transformer-style encoder over the candidate regulator set. A target-derived
\([{\rm CLS}]\)-style token is concatenated with regulator embeddings, processed
by self-attention without positional encodings, and mapped to a scalar residual
score, making the correction permutation-invariant over regulators. The residual
head is zero-initialized so training starts from the PairS2 backbone.

Residual HOS2 is trained with a margin-ranking loss that scores the true
regulator set above random and near-miss negative sets. For Residual HOS2, we use
exhaustive size-\(R\) subset decoding within the capped candidate pool when
feasible, and PairS2 proposal followed by HOS2 reranking for scalability.
Additional details are provided in Appendix~\ref{app:experimental-protocols}.

\subsection{BRIDGE Intervention Policy}

BRIDGE uses TRACE diagnoses to guide intervention selection. If retrieval is
the primary bottleneck, the candidate pool must be improved. If set-level
scoring is the primary bottleneck, interventions such as hard-negative
curricula or model-mined confuser repair can be evaluated. If decoding cost is
the primary bottleneck, scalable search procedures such as PairS2 proposal
followed by HOS2 reranking can be applied. BRIDGE accepts an intervention only
when it improves or safely preserves held-out recovery; otherwise, the
corresponding baseline pipeline component is retained.

\subsection{TRACE-lite for External Edge-Ranking Baselines}

TRACE-lite is the output-level branch of TRACE for external edge-ranking
baselines that do not expose explicit retrieval, set-level scoring, and
decoding modules. TRACE-lite converts each target-specific regulator ranking
into a fixed top-\(M\) candidate pool and a top-\(R\) predicted regulator set.
Operational retrieval failure means that at least one true regulator is absent
from the top-\(M\) pool. Operational set-selection failure means that the true
set is present in the top-\(M\) pool but is not recovered by the induced
top-\(R\) prediction. These terms describe output-level behavior only and
should not be interpreted as claims about the internal architecture or
cooperative modeling capacity of the external baseline. Additional TRACE-lite
details are provided in Appendix~\ref{app:trace_lite_foundation_models}.

\section{Experimental Setup}
We evaluate BRIDGE using controlled cooperativity benchmarks, SERGIO DS3, BEELINE Curated GSD, and mESC ChIP.
\subsection{Controlled Cooperativity Benchmark}
\label{sec:controlled-cooperativity}

We use a cooperativity-controlled synthetic benchmark as a TRACE stress-test
setting within BRIDGE. The purpose is to create controlled recovery tasks in which cooperative target generation can be systematically varied. The cooperativity parameter
\(C \in \{0.0,0.2,0.4,0.6,0.8,1.0\}\) controls the fraction of targets generated
by non-additive cooperative mechanisms. Thus, \(C=0.0\) corresponds to a fully
additive setting, while \(C=1.0\) corresponds to a fully cooperative setting.

A central concern in controlled cooperativity experiments is avoiding
feature--mechanism circularity. To avoid this shortcut, we use a leak-free
mechanism-mismatch benchmark: targets are generated by target-specific random
nonlinear MLP mechanisms, and Residual HOS2 operates directly on raw expression
vectors without handcrafted product-correlation features.

Each synthetic system contains \(320\) genes, consisting of \(80\) candidate
regulators and \(240\) target genes. Each target has a ground-truth regulator
set of size \(R=3\). We generate \(2000\) samples per synthetic system and use
noise standard deviation \(0.25\). Candidate pools are oracle-covered with the three true regulators and \(27\) distractors, giving \(M=30\). This controls retrieval
and isolates the set-level scoring problem. During exhaustive decoding, all
\(\binom{30}{3}=4060\) candidate regulator sets are scored for each target.

We compare decomposable PairS2 against Residual HOS2 across five random seeds. 
PairS2 aggregates independent
regulator--target evidence, whereas Residual HOS2 learns a set-level correction
on top of pairwise evidence. Because candidate pools are oracle-covered, this
experiment should be interpreted as a scoring-isolation benchmark rather than
an end-to-end retrieval experiment. The same cooperativity setting also supports model-agnostic TRACE-lite
evaluation of external edge-ranking baselines. Full implementation details are provided in
Appendix~\ref{app:cooperativity-simulation}.

Figure~\ref{fig:controlled-cooperativity-design} summarizes the resulting
stress-test design. The key point is that cooperativity is varied in the data
generation process, while retrieval is deliberately controlled by oracle-covered
candidate pools. This separates the question of whether the correct regulators
are available from the harder question of whether the scorer can rank the
complete regulator set above plausible distractor sets. The benchmark therefore
acts as a scoring-isolation test: PairS2 tests whether decomposable
regulator--target evidence is sufficient, whereas Residual HOS2 tests whether a
learned non-additive set correction improves recovery when the target mechanism
contains higher-order structure.
\FloatBarrier
\begin{figure}[t]
    \centering
    \includegraphics[width=\linewidth]{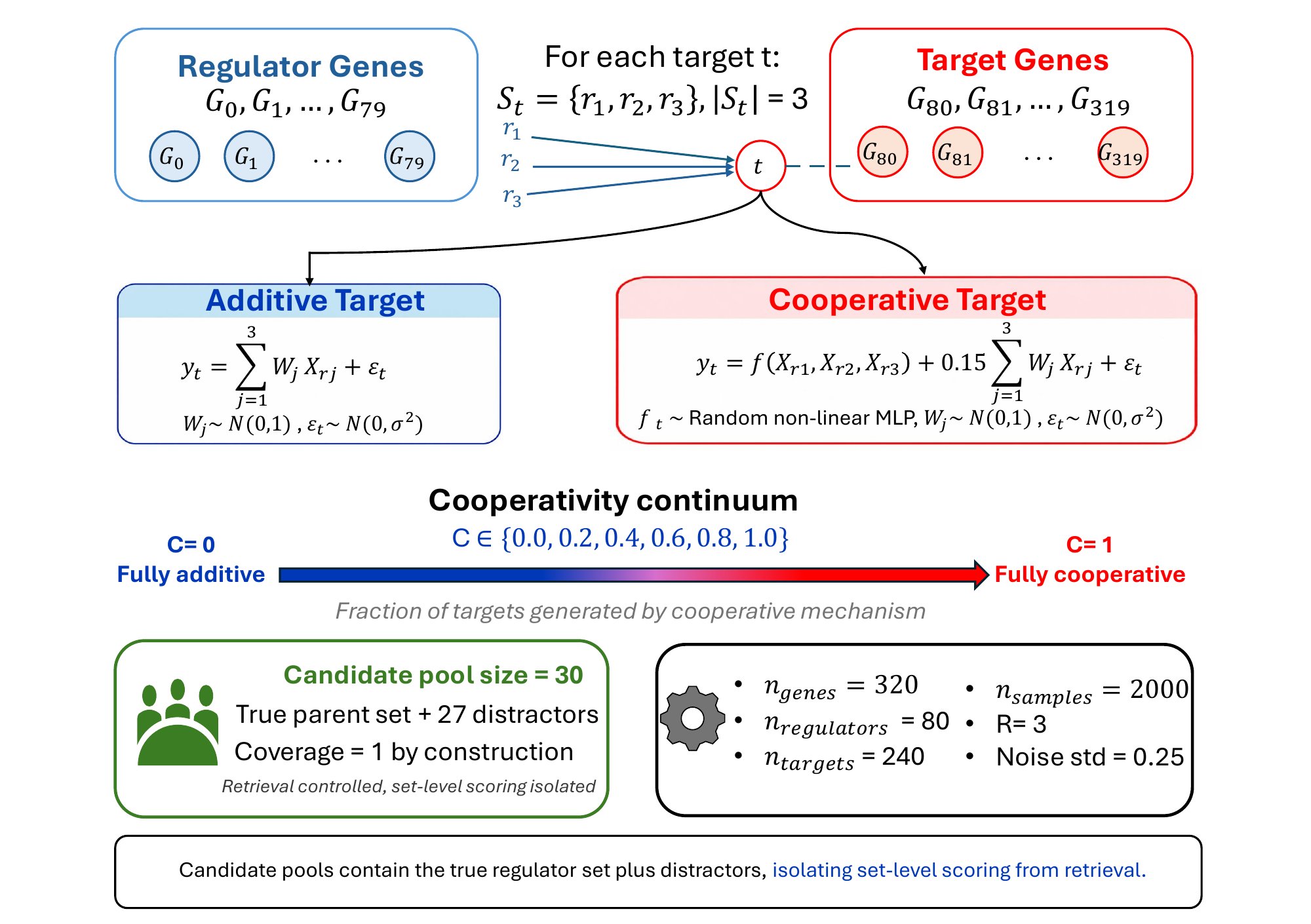}
    \caption{
     Corrected leak-free mechanism-mismatch cooperativity stress test. Regulator
     sets are fixed at \(R=3\). Additive targets are generated from linear regulator
     effects, whereas cooperative targets are generated by target-specific random
     nonlinear MLP mechanisms rather than product interactions. Candidate pools
     contain the true regulator set plus distractors, giving coverage equal to one
     by construction and isolating set-level scoring from retrieval.
     }
    \label{fig:controlled-cooperativity-design}
\end{figure}
\FloatBarrier

\subsection{SERGIO DS3 Benchmark}

We use SERGIO DS3 for a TRACE bottleneck analysis in a GRN-guided single-cell
benchmark with known regulatory structure \cite{Dibaeinia2020SERGIO}. We use the official
denoised DS3 setting with \(1200\) genes, \(9\) cell types, and \(300\) cells
per type. The ground-truth GRN is used to construct regulator-set recovery
tasks for \(R \in \{2,3,4\}\). Unless otherwise stated, we use decoding caps
\(M_R=\{80,80,55\}\) for \(R=\{2,3,4\}\). 

\subsection{Real-Data Stage-1 Retrieval Validation}

On the real mESC ChIP-seq regulatory network, the context-aware attention
retriever substantially improves strict regulator-set coverage over the pairwise
dot-product retriever. At \(K=200\), strict coverage increases from \(0.6873\) to
\(0.9893\), while edge recall increases from \(0.9900\) to \(0.9998\). This supports the practical relevance of BRIDGE's candidate-pool construction
stage on real data. Because mESC ChIP-derived regulator sets are large, we use
this experiment as real-data Stage-1 retrieval validation rather than claiming
full Residual HOS2 Stage-2 exact recovery.
Full results are provided in Appendix~\ref{app:mesc-stage1}.

\subsection{Evaluation Protocol}

We evaluate whether learned set-level scoring improves recovery under mechanism mismatch, whether retrieval alone explains exact-recovery failure, whether TRACE attributes failures to specific bottlenecks, whether interventions safely improve recovery, and whether external edge-ranking baselines recover complete regulator sets under TRACE-lite. For SERGIO DS3, we evaluate targets with regulator-set sizes
\(R\in\{2,3,4\}\) using candidate caps \(M_R=\{80,80,55\}\). We report Coverage@\(M\), exact recovery, conditional exact recovery,
Jaccard similarity, score gap, true-set rank, runtime, and candidate-set counts.
Unless otherwise stated, exact recovery denotes unconditional exact recovery
over all evaluated targets; conditional exact recovery is reported only when
restricted to targets whose true regulator set is present in the candidate pool.
Detailed experimental protocols are provided in
Appendix~\ref{app:experimental-protocols}.

\section{Results}
A multi-seed baseline re-evaluation is provided in Appendix~\ref{app:experimental-protocols}. This re-evaluation audits the two-stage recovery pipeline under the BRIDGE/TRACE diagnostic protocol and confirms that high candidate coverage alone does not yield high exact recovery.

Paired seed-blocked analyses found no significant Residual HOS2 gain
in Jaccard, recall, or exact recovery after Holm correction
(Appendix~\ref{app:sergio-pairs2-hos2-statistics}).

\subsection{Candidate-Cap and Runtime Sensitivity}

Increasing the candidate cap \(M\) improves Coverage@\(M\) but does not
monotonically improve unconditional exact recovery. For example, for \(R=3\), increasing
\(M\) from 55 to 80 raises coverage from 0.821 to 0.949, while unconditional exact recovery
remains 0.333. Full candidate-cap diagnostics are provided in
Appendix~\ref{app:candidate-cap-details}.

\subsection{Oracle Retrieval and Scoring Ceiling}

\begin{figure}[t]
\centering

\includegraphics[
    width=0.62\linewidth
]{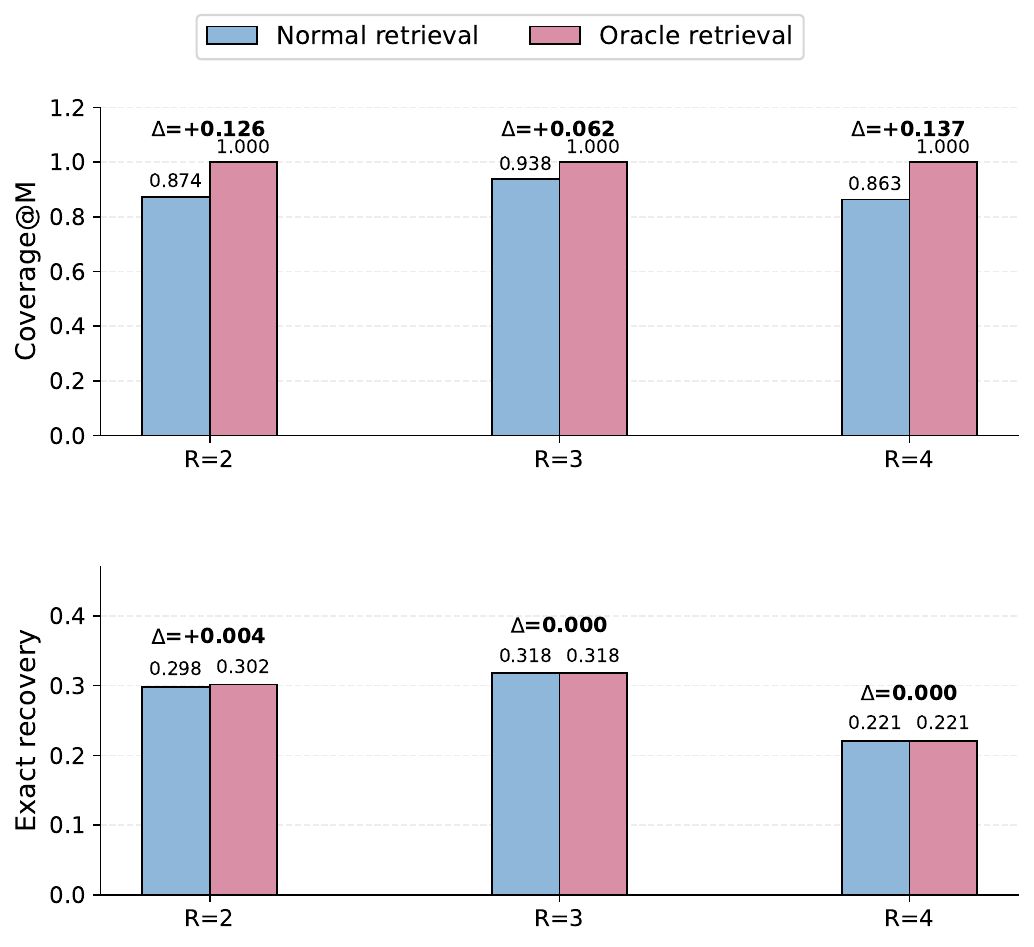}

\caption{
Five-seed oracle retrieval diagnostic on SERGIO DS3.
(a) Candidate coverage under normal and oracle retrieval.
(b) Exact recovery under normal and oracle retrieval.
Oracle retrieval raises Coverage@\(M\) to \(1.0\) for all regulator-set
sizes, but exact recovery changes only minimally. This indicates that
candidate absence is not the dominant remaining failure mode; set-level
scoring ambiguity is the primary bottleneck.
}
\label{fig:oracle-retrieval-comparison}

\vspace{0.7em}

\begin{minipage}{\linewidth}
\centering

\captionof{table}{
Paired effect of oracle retrieval relative to standard retrieval for
Residual HOS2 on SERGIO DS3. Positive values favor oracle retrieval.
}
\label{tab:oracle-retrieval-inference}

\small
\setlength{\tabcolsep}{5pt}
\renewcommand{\arraystretch}{1.12}

\begin{tabular}{lccc}
\toprule
Outcome
& Mean gain
& Blocked 95\% CI
& Holm $p$ \\
\midrule
Jaccard
& $+0.0017$
& $[-0.0014,\ 0.0048]$
& $.714$ \\
Recall
& $+0.0017$
& $[-0.0014,\ 0.0049]$
& $.714$ \\
Exact recovery
& $+0.0012$
& $[-0.0015,\ 0.0039]$
& $.714$ \\
\bottomrule
\end{tabular}

\vspace{1mm}

\parbox{\linewidth}{
\footnotesize
\textit{Notes.} Effects were estimated across 15 matched
seed--$R$ blocks. Oracle coverage equals one by construction.
Seed-profile bootstrap intervals also included zero for all outcomes.
Across 575 paired targets, oracle retrieval produced one additional
exact recovery and no regressions.
}

\end{minipage}
\end{figure}

Oracle retrieval raises Coverage@\(M\) to \(1.0\) for all regulator-set sizes
across five seeds, but unconditional exact recovery changes only minimally. For
\(R=2\), baseline unconditional exact recovery is \(0.298\pm0.066\), while
oracle retrieval gives \(0.302\pm0.064\). For \(R=3\), unconditional exact
recovery remains \(0.318\pm0.050\), and for \(R=4\), it remains
\(0.221\pm0.058\). Thus, forcing the complete true regulator set into every
candidate pool does not substantially improve unconditional exact recovery.
Because exhaustive decoding is used, the remaining failures indicate set-level
misranking rather than candidate absence. Additional oracle and score-gap diagnostics are provided in
Appendix~\ref{app:oracle-retrieval-ceiling} and
Appendix~\ref{app:oracle_scoregap_rank}.
Figure~\ref{fig:oracle-retrieval-comparison} turns the oracle setting into
a direct scoring-ceiling test. If exact recovery were mainly limited by
candidate absence, forcing every true regulator set into the candidate pool
would produce a large increase in exact recovery. Instead, oracle retrieval
mainly closes the retrieval gap while leaving exact recovery nearly unchanged.
Thus, the true set is often available to the decoder but still loses to a
higher-scoring near-miss set. This supports the TRACE interpretation that high
Coverage@\(M\) should be treated as a necessary retrieval condition, not as
evidence that cooperative regulator-set recovery is solved.
\FloatBarrier
Paired inference found no significant oracle-retrieval gain in Jaccard,
recall, or exact recovery after Holm correction. Thus, eliminating
candidate absence did not materially improve recovery, supporting
set-level scoring as the dominant remaining bottleneck. Additional
size-specific and sensitivity analyses are provided in
Appendix~\ref{app:oracle-retrieval-ceiling}.

\subsection{Recoverability Across Cooperativity Levels}
We next evaluate the corrected leak-free mechanism-mismatch cooperativity stress
test. This TRACE stress test varies the cooperativity parameter \(C\), while
removing product-based target-generation mechanisms and handcrafted
product-correlation features. Candidate pools are oracle-covered, so retrieval
is controlled and the experiment isolates set-level scoring rather than
end-to-end recovery.

\begin{figure}[!htbp]
\centering
\includegraphics[
    width=0.78\linewidth
]{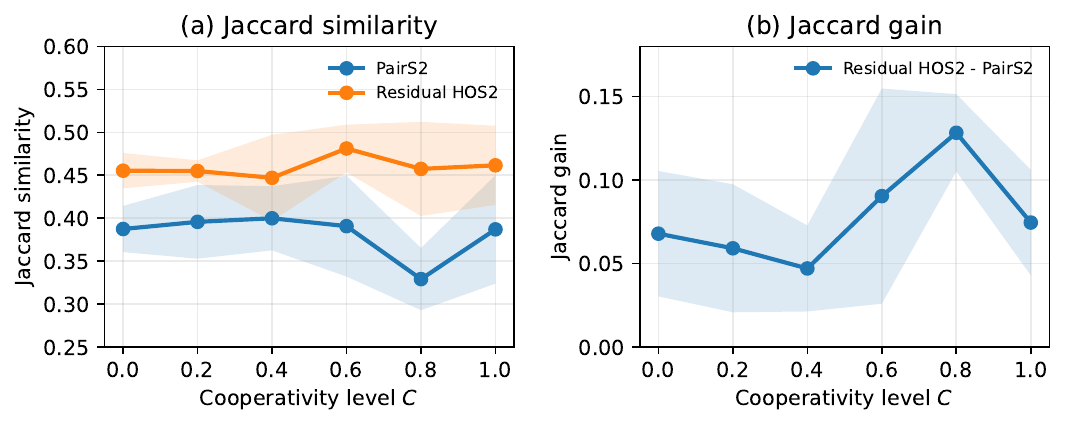}
\caption{
Leak-free mechanism-mismatch cooperativity stress test at \(R=3\).
(a) Jaccard similarity across cooperativity levels.
(b) Paired Jaccard gain of Residual HOS2 over PairS2.
Candidate pools are oracle-covered to isolate set-level scoring.
Lines show five-seed means; shaded bands show \(\pm 1\) standard deviation.
}
\label{fig:cooperativity-benchmark}
\end{figure}
\FloatBarrier
Across five seeds and six cooperativity levels, Residual HOS2 consistently
improves graded regulator-set recovery over decomposable PairS2. Averaged over
all \(30\) seed--cooperativity settings, PairS2 achieves Jaccard \(0.382\) and
recall \(0.522\), whereas Residual HOS2 achieves Jaccard \(0.460\) and recall
\(0.597\). Exact recovery also improves from \(0.053\) to \(0.113\), but remains
low overall.

The paired gains are $0.078 \pm 0.045$ in Jaccard,
$0.074 \pm 0.043$ in recall, and $0.060 \pm 0.050$ in exact
recovery. Residual HOS2 improves Jaccard and recall in all 30 matched
seed--cooperativity comparisons. Paired analyses blocking on seed and
cooperativity level confirmed significant improvements in Jaccard,
recall, and exact recovery after Holm correction, with seed-cluster
bootstrap confidence intervals excluding zero for all three outcomes
(Appendix~\ref{app:controlled-cooperativity-statistics}).
However, the Jaccard gain is not monotonic in $C$. Thus, the corrected benchmark supports a
weaker but more robust conclusion: learned set-level scoring improves graded
recovery under mechanism mismatch, while exact complete-set recovery remains a
strict bottleneck.

\subsection{Intervention Selection: Scoring Repair and Scalable Decoding}

\begin{figure}[t]
\centering
\includegraphics[width=0.62\linewidth]
{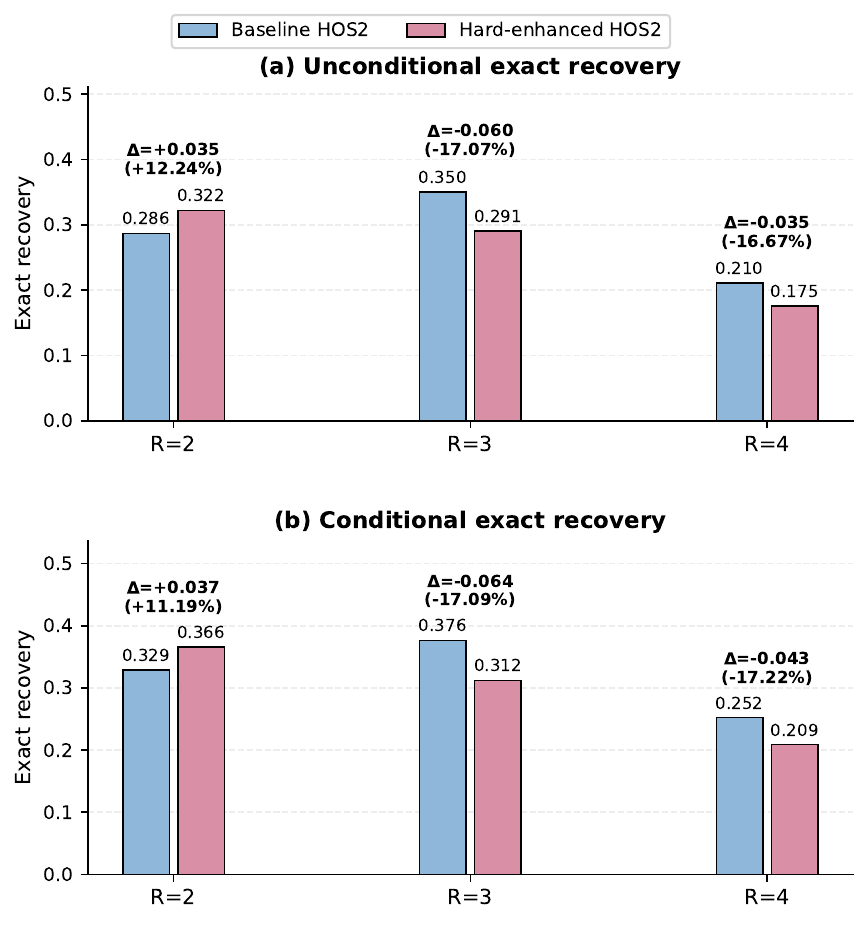}
\caption{
Effect of the hard-negative-enhanced curriculum on exact regulator-set recovery.
(a) Unconditional exact recovery.
(b) Conditional exact recovery among targets whose true regulator set is present
in the candidate pool.
Stronger near-miss hard-negative pressure improves recovery for \(R=2\) but
degrades recovery for \(R=3\) and \(R=4\), indicating that scoring intervention
is size-dependent.
}
\label{fig:hard-negative-ablation}
\end{figure}

Hard-negative repair is size-dependent
(Figure~\ref{fig:hard-negative-ablation};
Table~\ref{tab:hard-negative-inference}). The three-seed mean
unconditional exact recovery increases for $R=2$, from $0.2865$ to
$0.3216$, but decreases for $R=3$, from $0.3504$ to $0.2906$, and for
$R=4$, from $0.2105$ to $0.1754$. Exploratory paired analyses found no
significant overall gain after Holm correction. Thus, identifying a
scoring bottleneck does not imply that stronger hard-negative training
is uniformly safe. Full seed- and size-specific results are provided in
Appendix~\ref{app:hard-negative-intervention}.

This result illustrates why TRACE diagnoses should be treated as intervention
guidance rather than automatic prescriptions. A large scoring loss indicates
that the learned scorer often ranks a near-miss set above the true regulator
set, but it does not imply that every stronger scoring intervention will improve
held-out recovery. Hard-negative repair changes the scorer itself and can
reshape the ranking landscape differently across regulator-set sizes. The
\(R=2\) improvement suggests that confuser-focused training can help when the
near-miss space is relatively small, whereas the \(R=3\) and \(R=4\) drops show
that the same intervention can overcorrect in larger combinatorial regimes.
We also evaluate whether exhaustive HOS2 decoding can be made more scalable
without changing the learned HOS2 scorer. Exhaustive decoding evaluates all
size-\(R\) subsets inside the capped candidate pool, removing approximate search
as a confounder, but the number of evaluated sets grows combinatorially with
\(M\) and \(R\). PairS2 proposal followed by HOS2 reranking addresses this by
using the cheaper PairS2 scorer to propose a smaller regulator pool, then
applying HOS2 only within that reduced subset space.

Table~\ref{tab:decoder-scalability-5seed} reports the five-seed
descriptive comparison between exhaustive HOS2 decoding and PairS2
proposal followed by Residual HOS2 reranking.

\begin{table}[!htbp]
\centering
\small
\setlength{\tabcolsep}{5pt}
\renewcommand{\arraystretch}{1.18}
\caption{
Five-seed decoder scalability comparison between exhaustive HOS2
decoding and PairS2 proposal followed by HOS2 reranking.
}
\label{tab:decoder-scalability-5seed}

\begin{tabular}{@{}cccccc@{}}
\toprule
\(R\)
& \shortstack{Exh.\\Exact}
& \shortstack{Prop.\\Exact}
& \shortstack{\(\Delta\)\\Exact}
& \shortstack{Prop./Exh.\\Sets}
& \shortstack{Search\\Red.} \\
\midrule
2 
& \shortstack{0.2982\\\(\pm 0.0656\)}
& \shortstack{0.2947\\\(\pm 0.0625\)}
& \(-0.0035\)
& 190 / 3,160
& 93.99\% \\
\midrule
3
& \shortstack{0.3179\\\(\pm 0.0500\)}
& \shortstack{0.3333\\\(\pm 0.0480\)}
& \(+0.0154\)
& 2,300 / 82,160
& 97.20\% \\
\midrule
4
& \shortstack{0.2211\\\(\pm 0.0577\)}
& \shortstack{0.2211\\\(\pm 0.0577\)}
& 0.0000
& 12,650 / 341,055
& 96.29\% \\
\bottomrule
\end{tabular}
\end{table}
\FloatBarrier
\begin{table}[!htbp]
\centering
\caption{
Paired inference for PairS2 proposal followed by Residual HOS2
reranking relative to exhaustive Residual HOS2 decoding on SERGIO
DS3. Positive values favor proposal reranking.
}
\label{tab:decoder-scalability-inference}
\small
\setlength{\tabcolsep}{4.5pt}
\renewcommand{\arraystretch}{1.12}
\begin{tabular}{lccc}
\toprule
Outcome & Mean change & Blocked 95\% CI & Holm $p$ \\
\midrule
Jaccard & $+0.013$ & $[0.001,\ 0.024]$ & $.064$ \\
Recall & $+0.017$ & $[0.005,\ 0.029]$ & $.038$ \\
Exact recovery & $+0.004$ & $[-0.006,\ 0.013]$ & $.365$ \\
\bottomrule
\end{tabular}
\vspace{1mm}
\parbox{\columnwidth}{
\footnotesize
\textit{Notes.} Effects were estimated across 15 matched
seed--$R$ blocks using seeds 42--46. Confidence intervals are
unadjusted; $p$-values are Holm-adjusted across Jaccard, recall,
and exact recovery. Search-space reductions are deterministic
given the exhaustive and proposal-pool sizes and are therefore
reported descriptively. Failure to reject a difference is not,
by itself, a formal equivalence result.
}
\end{table}
\FloatBarrier

Figure~\ref{fig:hard-negative-ablation} shows that increasing
near-miss hard-negative pressure does not yield a uniform recovery
benefit across regulator-set sizes. For $R=2$, the three-seed mean
unconditional exact recovery increases from $0.2865$ to $0.3216$, and
conditional exact recovery increases from $0.3292$ to $0.3660$. In
contrast, both unconditional and conditional recovery decrease for
$R=3$ and $R=4$. Exploratory paired inference across nine matched
seed--$R$ blocks found no significant overall improvement in Jaccard,
recall, or exact recovery after Holm correction, and none of the
size-specific comparisons was significant. Across 345 paired targets,
the intervention produced 29 exact-recovery improvements and 32
regressions. These results show that identifying a scoring bottleneck
does not by itself justify stronger hard-negative training; scoring
repairs must be evaluated on held-out data and rejected when they
overcorrect or degrade recovery.

Paired inference across the 15 matched seed--$R$ blocks found a small
increase in graded recovery under proposal reranking. Recall increased
by $0.0169$ and remained significant after Holm correction
($p_{\mathrm{Holm}}=.038$). Jaccard increased by $0.0127$, but the
effect did not remain significant after multiplicity correction
($p_{\mathrm{Holm}}=.064$). Exact recovery changed by only $0.0040$
and was not significantly different from exhaustive decoding
($p_{\mathrm{Holm}}=.365$). The decoder effect did not vary
significantly across regulator-set sizes.

Across 575 paired targets, proposal reranking produced nine helpful
exact-recovery flips and seven harmful flips, leaving the exact outcome
unchanged for 559 targets. Together with the
$93.99\%$--$97.20\%$ reduction in HOS2-scored candidate sets, these
results support proposal reranking as a practical scalability
intervention under the evaluated proposal sizes. Because
nonsignificance is not a formal equivalence result, proposal coverage
should remain a monitored TRACE diagnostic.

\FloatBarrier

\subsection{TRACE Failure Attribution}
We next apply TRACE to decompose failed Residual HOS2 predictions on SERGIO DS3 into retrieval, set-level scoring, and decoding bottlenecks across seeds \(\{42,43,44,45,46\}\). Retrieval loss is defined as \(1-\mathrm{Coverage}@M\).

Under exhaustive decoding within the capped candidate pool, scoring loss is defined as
\[
\mathrm{ScoringLoss}
=
\mathrm{Coverage}@M-\mathrm{UnconditionalExact}.
\]
This quantity measures the fraction of targets for which the true regulator set is present in the candidate pool but is not selected as the final prediction.

Across five seeds, TRACE identifies set-level scoring as the dominant bottleneck for all regulator-set sizes. For \(R=2\), retrieval loss is \(0.126\pm0.040\), while scoring loss is substantially larger at \(0.575\pm0.079\). For \(R=3\), retrieval loss is \(0.062\pm0.014\), while scoring loss remains high at \(0.621\pm0.061\). For \(R=4\), retrieval loss is \(0.137\pm0.071\), and scoring loss is \(0.642\pm0.086\). Because decoding loss is zero under exhaustive enumeration, these results indicate that most remaining exact-recovery failures occur after the true regulator set has entered the candidate pool. The dominant failure mode is therefore set-level misranking rather than candidate absence or approximate search failure. Additional five-seed failure-attribution values are provided in
Appendix~\ref{app:trace-failure-attribution}.
\begin{figure}[H]
\centering
\includegraphics[width=0.95\linewidth]{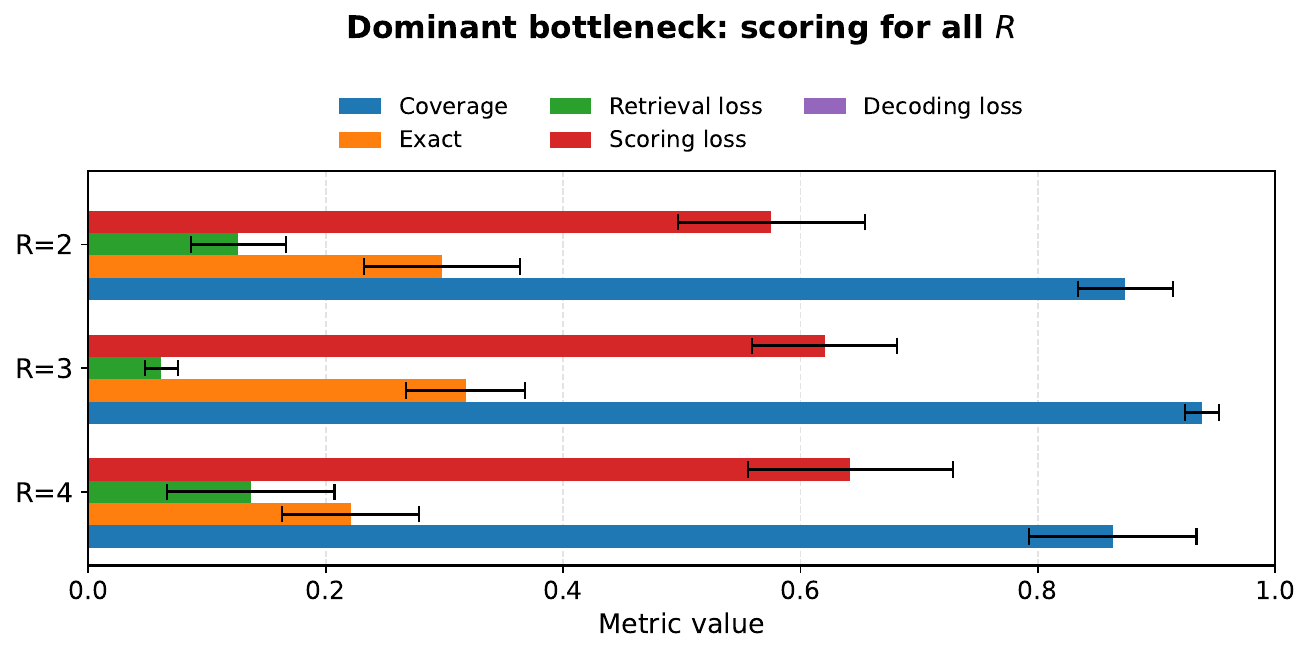}
\caption{TRACE failure attribution for Residual HOS2 on SERGIO DS3. Bars show five-seed mean values across seeds \(42\)--\(46\), with error bars denoting one standard deviation. Retrieval loss is \(1-\mathrm{Coverage}@M\), scoring loss is \(\mathrm{Coverage}@M-\mathrm{UnconditionalExact}\), and decoding loss is zero because exhaustive enumeration is used within the capped candidate pool. Scoring loss dominates retrieval loss for all regulator-set sizes.}
\label{fig:trace-attribution-bar}
\end{figure}
\FloatBarrier
Table~\ref{tab:trace-failure-attribution-five-seed} provides the
descriptive TRACE decomposition across the five seeds. Scoring loss
is larger than retrieval loss for every regulator-set size, while
decoding loss is zero because exhaustive enumeration is used within
the capped candidate pool.

To formally test the dominant-bottleneck conclusion, we define the
matched seed--\(R\) contrast
\[
\Delta_{\mathrm{TRACE}}
=
\mathrm{ScoringLoss}
-
\mathrm{RetrievalLoss}.
\]
Positive values indicate that set-level scoring contributes more
failure than candidate retrieval.

Across the 15 matched seed--\(R\) blocks, scoring loss exceeded
retrieval loss by \(0.504\) on average
(blocked 95\% CI \([0.433,0.576]\), \(p<.001\)).
The seed-profile bootstrap 95\% confidence interval
\([0.462,0.549]\) also excluded zero. Thus, the conclusion that
set-level scoring is the dominant remaining bottleneck is supported
by formal inference rather than only by the descriptive mean
decomposition.

Exploratory regulator-set-size-specific contrasts were positive for
all three sizes and remained significant after Holm correction.
The mean difference was \(0.449\) for \(R=2\)
(\(p_{\mathrm{Holm}}=.001\)), \(0.559\) for \(R=3\)
(\(p_{\mathrm{Holm}}<.001\)), and \(0.505\) for \(R=4\)
(\(p_{\mathrm{Holm}}=.002\)). The contrast did not vary significantly
across regulator-set sizes, \(F(2,8)=1.042\), \(p=.396\), indicating
that scoring dominance was consistent across the evaluated
combinatorial regimes.

Scoring was the largest TRACE loss in all 15 seed--\(R\) blocks.
The exact five-seed profile sign-flip sensitivity test gave
\(p=.0625\), the smallest attainable two-sided value with five
complete seed profiles. This finite-seed result reflects the limited
resolution of the exact sensitivity test; the blocked confidence
interval and seed-profile bootstrap both provide direct interval
support for a positive scoring-minus-retrieval contrast.

\begin{table}[t]
\centering
\caption{
Formal TRACE failure-attribution inference for Residual HOS2 on
SERGIO DS3. The primary contrast is scoring loss minus retrieval loss;
positive values indicate a larger scoring bottleneck.
}
\label{tab:trace-failure-attribution-inference}
\small
\setlength{\tabcolsep}{4.2pt}
\renewcommand{\arraystretch}{1.10}
\begin{tabular}{lccc}
\toprule
Comparison & Mean difference & 95\% CI & $p$ \\
\midrule
Overall blocked & $0.504$ & $[0.433,\ 0.576]$ & $<.001$ \\
$R=2$ & $0.449$ & $[0.317,\ 0.581]$ & $.001$ \\
$R=3$ & $0.559$ & $[0.468,\ 0.650]$ & $<.001$ \\
$R=4$ & $0.505$ & $[0.323,\ 0.688]$ & $.002$ \\
\bottomrule
\end{tabular}

\vspace{1mm}
\parbox{\columnwidth}{
\footnotesize
\textit{Notes.} The overall analysis uses 15 matched seed--$R$
blocks from seeds 42--46. The blocked confidence interval uses a
two-way additive model with seed and regulator-set-size effects.
The seed-profile bootstrap 95\% interval was
$[0.462,\
\ 0.549]$.
The exact five-seed profile sign-flip sensitivity test gave
$p=0.0625$; with five seed
profiles, this is the smallest attainable two-sided sign-flip
$p$-value. The $R$-specific $p$-values are Holm-adjusted across
the three regulator-set sizes. The test for variation across $R$
gave $F(2,
8)
=1.042$,
$p=.396$.
}
\end{table}
\FloatBarrier

\subsection{TRACE-lite Evaluation of External Edge-Ranking Baselines}

To evaluate whether standard edge-ranking GRN methods recover complete
cooperative regulator sets, we apply TRACE-lite to correlation, mutual
information, GRNBoost2, and GENIE3. Each method produces target-specific
regulator rankings, which are converted into a top-\(M\) candidate pool and a
top-\(R\) predicted regulator set for each target.

Table~\ref{tab:external-trace-lite-results} shows that external edge-ranking
baselines do not directly solve cooperative regulator-set recovery under a fixed
top-\(M\)/top-\(R\) TRACE-lite conversion protocol. For \(R=2\), correlation and
GENIE3 achieve relatively high candidate coverage, 62.0\% and 57.3\%,
respectively, but exact recovery remains low at 1.8\% and 1.2\%. GRNBoost2
achieves 50.3\% coverage and the highest \(R=2\) exact recovery among external
baselines at 2.9\%. For higher-order sets, all external baselines remain near
zero exact recovery. GENIE3 obtains 9.4\% coverage and 0.0\% exact recovery for
\(R=3\), and 3.5\% coverage and 0.0\% exact recovery for \(R=4\). These results
support the central TRACE-lite finding: strong edge rankings may retrieve some
true regulators, but edge-level ranking alone does not reliably recover complete
cooperative regulator sets.

TRACE-lite therefore separates operational retrieval failure from top-\(R\)
selection failure in external edge-ranking baselines.
\FloatBarrier
\begin{table}[H]
\centering
\footnotesize
\setlength{\tabcolsep}{3.2pt}
\renewcommand{\arraystretch}{1.05}
\caption{TRACE-lite evaluation of external edge-ranking baselines on SERGIO DS3.}
\label{tab:external-trace-lite-results}
\begin{tabular}{@{}lcccccc@{}}
\toprule
Method & \(R\) & \(M\) & Cov.@\(M\) & Exact & Cond. Exact & Jaccard \\
\midrule
Correlation & 2 & 80 & 0.620 & 0.018 & 0.028 & 0.216 \\
Correlation & 3 & 80 & 0.137 & 0.000 & 0.000 & 0.118 \\
Correlation & 4 & 55 & 0.053 & 0.000 & 0.000 & 0.057 \\
\midrule
MI & 2 & 80 & 0.345 & 0.023 & 0.068 & 0.113 \\
MI & 3 & 80 & 0.043 & 0.000 & 0.000 & 0.058 \\
MI & 4 & 55 & 0.000 & 0.000 & -- & 0.018 \\
\midrule
GRNBoost2 & 2 & 80 & 0.503 & 0.029 & 0.058 & 0.191 \\
GRNBoost2 & 3 & 80 & 0.094 & 0.000 & 0.000 & 0.100 \\
GRNBoost2 & 4 & 55 & 0.000 & 0.000 & -- & 0.050 \\
\midrule
GENIE3 & 2 & 80 & 0.573 & 0.012 & 0.020 & 0.181 \\
GENIE3 & 3 & 80 & 0.094 & 0.000 & 0.000 & 0.116 \\
GENIE3 & 4 & 55 & 0.035 & 0.000 & 0.000 & 0.064 \\
\bottomrule
\end{tabular}
\end{table}
\FloatBarrier

\subsection{Real Gold-Standard TRACE-lite Evaluation}
To validate TRACE-lite beyond simulations, we evaluated real curated
gold-standard GRN labels from BEELINE Curated GSD. Incoming gold-standard
regulator--target edges were grouped by target to define target-specific
regulator sets. After restricting all methods to the shared
foundation-model-mapped gene universe and removing self-edges, the evaluation
contained 30 GSD variants and 450 target instances with mean set size \(3.87\).
\FloatBarrier
\begin{table}[H]
\centering
\footnotesize
\setlength{\tabcolsep}{3.2pt}
\renewcommand{\arraystretch}{1.05}
\caption{TRACE-lite on BEELINE Curated GSD gold-standard regulator sets.}
\label{tab:beeline-gsd-gold-standard}
\begin{tabular}{@{}lcccc@{}}
\toprule
Method & Cov.@10 & Exact & Jac. & Rec. \\
\midrule
Pearson & 0.518 & 0.000 & 0.254 & 0.379 \\
MI & 0.502 & 0.000 & 0.245 & 0.366 \\
GENIE3 & 0.540 & 0.000 & 0.254 & 0.380 \\
GRNBoost2 & 0.504 & 0.002 & 0.254 & 0.379 \\
Geneformer Probe & 0.267 & 0.000 & 0.163 & 0.263 \\
scGPT Probe& 0.267 & 0.000 & 0.174 & 0.264 \\
\bottomrule
\end{tabular}
\vspace{-0.6em}
\end{table}
\FloatBarrier

Table~\ref{tab:beeline-gsd-gold-standard} shows that exact complete-set
recovery is nearly absent, while graded metrics reveal partial regulator-set
signal. GENIE3 gives the best Coverage@10 in this small curated benchmark, and
expression/tree-based baselines recover stronger signal than frozen FM
embedding probes. These FM results are interpreted as zero-shot representation
probes rather than directed GRN inference.
A zero-shot mESC retrieval probe further supports this pattern
(Appendix~\ref{app:mesc-stage1}): at \(K=200\), Geneformer and scGPT recover
many individual ChIP regulators (EdgeRec \(=0.864\) and \(0.859\)) but rarely
recover complete large regulator sets (StrictCov \(=0.026\) and \(0.028\)).

Together, BEELINE and mESC provide two complementary real-data checks. BEELINE
tests TRACE-lite on curated target-specific regulator labels, allowing external
rankers and frozen foundation-model probes to be compared under the same
top-\(M\)/top-\(R\) set-recovery protocol. mESC ChIP tests whether Stage-1
retrieval can place large real ChIP-derived regulator sets inside the candidate
pool. These real-data experiments therefore support the central distinction
between regulator-level recall and complete regulator-set recovery, while
leaving full Residual HOS2 Stage-2 exact recovery on large real regulator sets
as future work.

\section{Discussion}

BRIDGE reframes GRN inference as complete regulator-set recovery rather than only edge ranking. Across controlled simulations, SERGIO DS3, and real gold-standard evaluations, candidate coverage and regulator-level recall improve recoverability but do not guarantee exact complete-set recovery. Even when many true regulators are retrieved, the full regulator set can remain unrecovered because the model or baseline selects an incorrect near-miss set. The dominant remaining failure mode is therefore set-level recovery failure rather than simple candidate absence.

The corrected cooperativity stress test supports a more cautious conclusion.
After replacing product-based target-generation mechanisms and removing
handcrafted product-correlation features, Residual HOS2 no longer shows a
monotonic advantage as \(C\) increases. Instead, it consistently improves
graded recovery over PairS2 across all cooperativity levels, while exact
recovery remains low. This suggests that learned set-level scoring can improve
regulator-set overlap under mechanism mismatch, but it does not by itself solve
complete exact recovery. The stress test therefore supports the value of
set-level scoring while reinforcing the need for TRACE diagnostics.

TRACE provides a direct explanation for the remaining failure pattern. Across
regulator-set sizes \(R \in \{2,3,4\}\), scoring loss is consistently larger
than retrieval loss, while decoding loss is zero under exhaustive enumeration.
The dominant bottleneck is therefore set-level misranking: the model often
assigns a higher score to an incorrect near-miss set than to the true regulator
set. This explains why increasing candidate-pool coverage or forcing oracle
retrieval produces only limited gains in exact recovery.
This re-evaluation also refines the earlier workshop conclusion: the key
contribution is not that Residual HOS2 uniformly solves exact regulator-set
recovery, but that BRIDGE/TRACE exposes when recovery is limited by retrieval,
set-level scoring, decoding cost, or evaluation mismatch.

The operational bottleneck view also clarifies why exact regulator-set recovery
is stricter than either edge recovery or candidate coverage. Retrieval preserves
access to the true regulators, but it does not ensure that the true combination
is preferred over high-overlap alternatives. Scoring must therefore preserve
set-level information: the model must assign the complete true set a higher
score than plausible near-miss sets that share some, but not all, true
regulators. Decoding then determines whether the highest-scoring set can be
found under combinatorial constraints. In this sense, BRIDGE/TRACE converts a
single recovery number into a staged diagnosis of where recovery-relevant
information is discarded, obscured, or made computationally inaccessible.

On SERGIO DS3, TRACE-lite shows that correlation, mutual information, GRNBoost2,
and GENIE3 do not directly solve cooperative regulator-set recovery: the best \(R=2\)
exact recovery is only 2.9\%, and all baselines reach 0.0\% exact recovery for
\(R=3\) and \(R=4\). Thus, edge-level ranking and cooperative set-level recovery
are distinct objectives.

The BEELINE GSD evaluation provides a preliminary real-data TRACE-lite test using curated target-specific regulator sets. Exact complete-set recovery is rare, while graded metrics show partial regulator-set signal. GENIE3 obtains the highest Coverage@10, and frozen Geneformer/scGPT probes remain weaker than expression- and tree-based baselines.

The intervention results show why bottleneck-aware validation is necessary.
Although TRACE identifies scoring as the dominant failure mode, stronger scoring
interventions are not uniformly beneficial. Hard-negative-enhanced training and
model-mined confuser fine-tuning improve \(R=2\) recovery but degrade larger
regulator sets, suggesting that direct confuser-based repair can overcorrect the
scorer in larger combinatorial regimes. BRIDGE therefore acts as an
intervention-selection safeguard: it accepts scoring interventions only when
they improve held-out recovery and rejects them when they degrade performance.

In contrast, proposal-based decoding provides a safer scalability intervention.
PairS2 proposal followed by HOS2 reranking largely preserves exact-recovery
behavior across five seeds while reducing the number of HOS2-scored candidate
sets by approximately \(94\)--\(97\%\). Thus, decoding is not the dominant
recovery bottleneck under exhaustive enumeration, but it remains a practical
scalability bottleneck that BRIDGE can address without changing the learned HOS2
scorer.

Overall, by separating retrieval, scoring, decoding, and evaluation, BRIDGE makes the distinction between edge-ranking success and cooperative set recovery measurable.

\section{Limitations and Future Directions}
\label{sec:limitations-future}
\label{sec:foundation-model-extensions}

This study evaluates cooperative regulator-set recovery in settings with known
ground truth. SERGIO DS3 and the controlled cooperativity stress test support
diagnosis of retrieval, scoring, decoding, and evaluation bottlenecks, but they
should be interpreted as recoverability benchmarks rather than direct biological
validation. The BEELINE GSD and mESC experiments provide preliminary real-data evaluations
using curated gold-standard regulator--target labels, but they do not establish
full Residual HOS2 exact recovery on large real regulator sets.
Moreover, this study assumes that the regulator-set size \(R\) is known during evaluation.
This is appropriate for controlled diagnostic benchmarks such as SERGIO DS3,
where targets can be stratified by ground-truth regulator-set size, but real
deployment would require selecting \(R\) per target using validation, calibrated
score thresholds, or stopping rules over candidate set sizes. Unknown-\(R\)
inference is left for future work.

A first future direction is to develop semi-synthetic real-expression
benchmarks that combine real single-cell expression with curated TF--target
annotations and known cooperative regulator-set labels. This would preserve
realistic expression structure while enabling exact set-level evaluation.

A second direction is to scale Stage-2 recovery beyond small regulator sets.
Because exhaustive HOS2 decoding becomes infeasible for large regulator
neighborhoods, future work should evaluate approximate decoders such as beam
search, local refinement, learned proposals, or PairS2-guided reranking, with
TRACE auditing any new decoding failures.

A third direction is to connect BRIDGE with single-cell foundation models.
Embeddings, perturbation-response scores, attention-derived gene associations,
or graph-enhanced representations could be tested within TRACE-lite to determine
whether they improve coverage, score separation, true-set rank, or exact
regulator-set recovery.

\section{Conclusion}

BRIDGE provides a model-agnostic framework for diagnosing cooperative
regulator-set recovery from gene expression data. Rather than treating gene
regulatory network inference as a single edge-ranking task, BRIDGE separates
candidate retrieval, set-level scoring, combinatorial decoding, and outcome
evaluation. TRACE turns these stages into measurable bottlenecks using coverage,
score gaps, true-set ranks, decoding comparisons, and exact recovery.

Across controlled cooperativity regimes, TRACE shows that learned set-level scoring improves graded recovery under mechanism mismatch, although exact complete-set recovery remains difficult. On SERGIO DS3, oracle retrieval and failure attribution identify set-level misranking—not candidate coverage—as the dominant remaining bottleneck. Proposal-based reranking also reduces decoding cost while largely preserving recovery behavior.

Real BEELINE GSD and mESC evaluations support the same broader conclusion:
recovering many individual regulators is not equivalent to recovering the
complete regulator set. Overall, BRIDGE/TRACE makes edge ranking,
candidate-pool coverage, graded regulator overlap, and exact cooperative
mechanism recovery measurable as distinct objectives.

The complete code, experimental configurations, and evaluation scripts will be made publicly available upon acceptance.
\clearpage
\bibliographystyle{plainnat}
\bibliography{references}
\clearpage
\appendix

\setcounter{secnumdepth}{2}
\setcounter{section}{0}
\renewcommand{\thesection}{\Alph{section}}
\renewcommand{\thesubsection}{\thesection.\arabic{subsection}}

\section{SERGIO DS3 Reproducibility}
\label{app:sergio-reproducibility}
\subsection{Official Repository and Version}

To ensure reproducibility, we reconstructed the SERGIO DS3 data source from the official SERGIO repository. The repository was cloned and the exact commit was recorded. The commit used in this study was \texttt{a6190b74425112834c8fa9b4b6157d9cb3d1ab88}.

The official DS3 data folder was \texttt{data\_sets/De-noised\_1200G\_9T\_300cPerT\_6\_DS3/}. This folder corresponds to the de-noised DS3 setting with 1200 genes, 9 cell types, and 300 cells per cell type.

\subsection{Files Verified}

The official DS3 folder contained 18 files: 15 expression replicate files,
one ground-truth GRN file, one regulator-profile file, and one
interaction-parameter file. The expression replicate files were named
\path{simulated_noNoise_0.csv} through
\path{simulated_noNoise_14.csv}. The ground-truth GRN file was
\path{gt_GRN.csv}. The regulator-profile file was
\path{Regs_cID_6.txt}. The interaction-parameter file was
\path{Interaction_cID_6.txt}.

The ground-truth GRN file \path{gt_GRN.csv} had shape
\(2713 \times 2\), corresponding to 2713 regulator--target interactions.
The regulator-profile file \path{Regs_cID_6.txt} had shape
\(67 \times 10\). The interaction file
\path{Interaction_cID_6.txt} was not a rectangular table; it was a ragged
comma-separated interaction-parameter file with variable-length rows.

\subsection{Ground-Truth Regulator-Set Size Distribution}

We further inspected the reproduced DS3 ground-truth GRN to determine the number of regulators associated with each target gene. The reproduced DS3 ground truth contained 1133 target genes with at least one regulator. Regulator-set sizes ranged from 1 to 11, with a mean of 2.39 regulators per target.

\begin{figure}[H]
\centering
\includegraphics[width=0.58\linewidth]{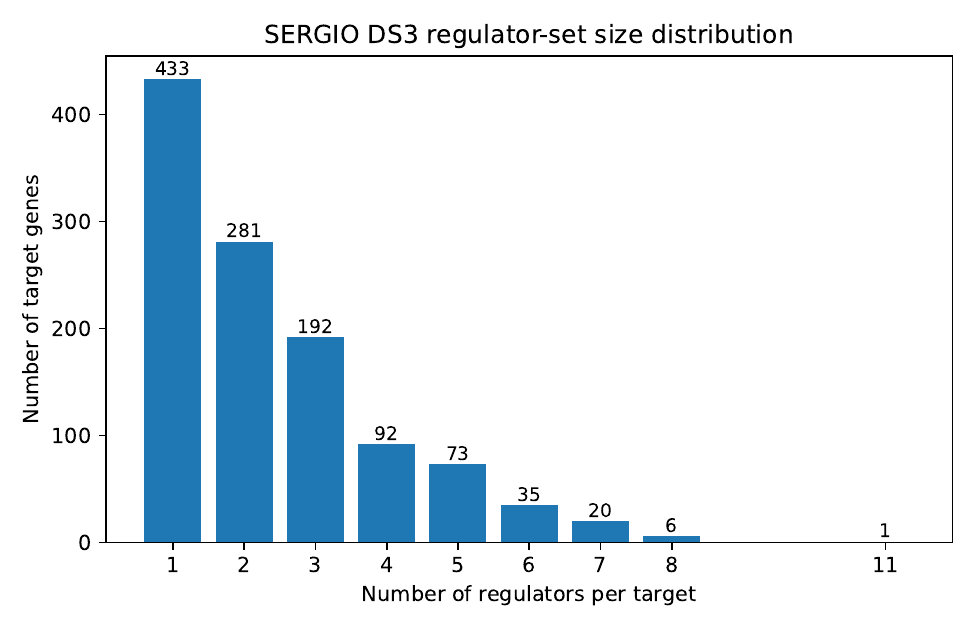}
\caption{Regulator-set size distribution in the reproduced SERGIO DS3 ground-truth GRN. The dataset contains 1133 regulated target genes, with regulator-set sizes ranging from 1 to 11. The fixed-size higher-order recovery experiments focus on $R \in \{2,3,4\}$, corresponding to 281, 192, and 92 targets, respectively.}
\label{fig:sergio-size-distribution}
\end{figure}

This distribution motivates stratifying SERGIO DS3 targets by regulator-set size before applying fixed-size higher-order recovery models. In particular, we use targets with $R \in \{2,3,4\}$ for HOS2-style regulator-set recovery experiments, while single-regulator targets primarily support pairwise recovery analysis.

\subsection{Expression Matrix Verification}

Each expression replicate file was initially stored with one label row and one label column. When loaded using the first column as the gene index, each replicate had shape $1200 \times 2700$. This corresponds to 1200 genes and 2700 cells, where 2700 cells arise from 9 cell types with 300 cells per type.

The 15 replicate matrices were concatenated along the cell dimension. The reproduced DS3 expression matrix therefore had shape $1200 \times 40500$, because $15 \times 2700 = 40500$.

\subsection{Preprocessing Rule}

Let $X_i$ denote the expression matrix from replicate $i$, where $i \in \{0,\ldots,14\}$. Each $X_i$ has shape $1200 \times 2700$. The reproduced DS3 matrix was constructed by concatenating the 15 matrices along the cell dimension:

\[
X_{\mathrm{DS3}} =
\left[
X_0,
X_1,
\ldots,
X_{14}
\right].
\]

No expression values were modified during this concatenation step. The operation only reorganized the official SERGIO replicate files into a single matrix format suitable for downstream PairS2, HOS2, and BRIDGE experiments.

\subsection{Role in This Study}

In addition to the cooperativity-controlled synthetic benchmark, we evaluate
BRIDGE on SERGIO-generated single-cell gene-expression data. SERGIO is a
GRN-guided single-cell simulator that generates expression profiles from a
specified regulatory network, making it suitable for benchmarking GRN
reconstruction methods \cite{Dibaeinia2020SERGIO}.

The SERGIO DS3 benchmark is used as a biologically motivated expression-only
benchmark with a known ground-truth GRN. It supports evaluation of retrieval,
scoring, decoding, and exact regulator-set recovery under a controlled simulated
setting. Because the standard SERGIO DS3 setting is not designed as a
cooperativity-controlled benchmark, we use DS3 primarily for baseline recovery,
bottleneck diagnosis, and external-baseline evaluation. Explicit
cooperativity-controlled experiments are conducted separately using synthetic
datasets in which the cooperativity parameter is directly varied.

\section{Real-Data Stage-1 Retrieval Validation on mESC}
\label{app:mesc-stage1}

We evaluate Stage-1 retrieval on a real gene regulatory network derived from an
mESC ChIP-seq network paired with a single-cell expression matrix. mESC has gold-standard regulator--target edges and therefore target-specific
gold-standard regulator sets, but the sets are too large for exhaustive HOS2
decoding. Therefore, we use mESC for Stage-1 retrieval validation rather than
full Stage-2 exact recovery. Although beam search could be used to approximate Stage-2 decoding on mESC, we
do not treat it as full exact regulator-set recovery in this work. The mESC
ChIP-derived regulator sets are large, with mean and maximum in-degree 65.7 and
182, making exact complete-set decoding qualitatively different from the
small-\(R\) SERGIO setting. Beam search would test proposal-based reranking over
large ChIP-derived sets, but would not guarantee recovery of the full
gold-standard set and would require a scorer trained and validated for this
large-set regime. We therefore use mESC as real-data Stage-1 retrieval
validation and leave beam-based large-set Stage-2 reranking for future work. The
expression matrix contains \(G=18{,}386\) genes and \(C=421\) cells, and the
candidate regulator set contains \(|\mathcal{R}|=229\) regulators. After
gene-name mapping and filtering, the directed network contains 806,646 kept
edges and 12,284 target genes, with mean, median, and maximum in-degree
65.7, 71, and 182, respectively. We split targets into 4,000 train targets and
1,500 test targets.

Because the regulator universe is known, we rank only candidate regulators for
each target. We report two retrieval metrics. Strict Coverage@\(K\) measures
whether the complete ChIP-derived regulator set is contained in the top-\(K\)
ranked regulators:
\[
\mathrm{StrictCov@}K
=
\mathbb{E}_{t}
\left[
\mathbb{I}
\left(
\mathcal{R}^{\star}(t)
\subseteq
\mathrm{Top}\text{-}K(t)
\right)
\right].
\]
Edge-Recall@\(K\) measures the average fraction of true regulators recovered
within the top-\(K\) list:
\[
\mathrm{EdgeRec@}K
=
\mathbb{E}_{t}
\left[
\frac{
|\mathcal{R}^{\star}(t)\cap \mathrm{Top}\text{-}K(t)|
}{
|\mathcal{R}^{\star}(t)|
}
\right].
\]

\begin{table}[H]
\centering
\footnotesize
\setlength{\tabcolsep}{3.0pt}
\renewcommand{\arraystretch}{1.12}
\caption{mESC Stage-1 retrieval validation on 1,500 real-data test targets. Strict Coverage@\(K\) requires the full regulator set to be included in the top-\(K\) ranked regulators. Edge-Recall@\(K\) is the average fraction of true regulators recovered within the top-\(K\) list.}
\label{tab:mesc-stage1}
\begin{tabular}{@{}ccccc@{}}
\toprule
\(K\)
& \shortstack{Pairwise\\StrictCov}
& \shortstack{Pairwise\\EdgeRec}
& \shortstack{Context-Attn\\StrictCov}
& \shortstack{Context-Attn\\EdgeRec} \\
\midrule
100 & 0.0627 & 0.8960 & 0.0727 & 0.9050 \\
120 & 0.1227 & 0.9385 & 0.1627 & 0.9510 \\
140 & 0.2633 & 0.9665 & 0.3980 & 0.9791 \\
160 & 0.4247 & 0.9780 & 0.7040 & 0.9933 \\
180 & 0.5567 & 0.9836 & 0.9047 & 0.9979 \\
200 & 0.6873 & 0.9900 & 0.9893 & 0.9998 \\
\bottomrule
\end{tabular}
\end{table}

Table~\ref{tab:mesc-stage1} shows that the context-aware attention retriever
substantially improves strict regulator-set coverage on real mESC data. At
\(K=200\), the attention retriever reaches 0.9893 Strict Coverage@\(K\),
whereas the pairwise dot-product retriever reaches 0.6873. This result supports
the use of context-aware retrieval as a practical candidate-pool construction
strategy beyond controlled simulated benchmarks.

This experiment should be interpreted as retrieval validation rather than full
exact regulator-set recovery. Unlike the SERGIO DS3 experiments, where
\(R \in \{2,3,4\}\) allows exhaustive subset decoding, real mESC targets have
large regulator sets. Full Stage-2 HOS2 decoding for these large sets would
require approximate search or a semi-synthetic real-expression benchmark with
known cooperative regulator-set labels.
\begin{table}[H]
\centering
\footnotesize
\setlength{\tabcolsep}{3.2pt}
\renewcommand{\arraystretch}{1.05}
\caption{Zero-shot FM retrieval probes on mESC at \(K=200\).}
\label{tab:mesc-fm-stage1}
\begin{tabular}{@{}lccccc@{}}
\toprule
Method & \(N\) & \(\bar{R}\) & SC & ER & ER-M \\
\midrule
Geneformer & 10920 & 67.08 & 0.026 & 0.864 & 0.903 \\
scGPT & 11002 & 66.84 & 0.028 & 0.859 & 0.897 \\
\bottomrule
\end{tabular}
\end{table}
Here, SC denotes Strict Coverage, ER denotes Edge Recall over the full
ChIP-derived regulator set, and ER-M denotes Edge Recall restricted to
mappable true regulators.

The zero-shot FM probes recover many individual ChIP-derived regulators by
\(K=200\), but rarely recover complete large regulator sets. This supports the
distinction between regulator-level recall and strict set coverage. The result
is interpreted cautiously because the benchmark is mouse mESC and the models
are used only through gene-symbol embedding similarity, not as directed GRN
inference methods.
\section{Background}

\subsection{Pairwise Regulatory Discovery}

Traditional gene regulatory network inference methods focus on recovering pairwise relationships:

\[
r_i \rightarrow t
\]

This formulation assumes regulators act independently.

\subsection{Cooperative Regulation}

Many biological systems exhibit cooperative regulation, where the expression of a target gene is not determined by a single regulator acting independently, but by the joint activity of multiple transcription factors or regulatory elements. In enhancer-mediated regulation, transcription factors often bind in combinations, and their regulatory effects depend on cellular, spatial, and molecular context. This motivates a higher-order view of gene regulation in which regulator sets, rather than isolated regulator-target edges, form the basic unit of mechanism discovery \cite{Spitz2012Transcription}.

Many biological systems exhibit cooperative behavior where target expression depends on combinations of regulators:

\[
\{A,B,C\} \rightarrow T
\]

In this example, regulators $A$, $B$, and $C$ jointly regulate the target gene $T$. In such systems, individual regulators may exhibit weak predictive signals despite the collective set being highly informative.

In this work, we define a cooperative regulatory mechanism as a target-specific relationship in which a set of regulators $S_t = \{r_1,r_2,\ldots,r_k\}$ jointly controls the expression of a target gene $t$ in a way that cannot be fully decomposed into independent pairwise regulator-target effects. In contrast to an additive model, where each regulator contributes separately, a cooperative mechanism may produce nonlinear, context-dependent, or threshold-like effects.

Formally, an additive regulatory model can be written as:

\[
f_t(S_t) \approx \sum_{r_i \in S_t} f_t(r_i)
\]

Here, the total regulatory effect is approximately the sum of individual regulator effects. In contrast, a cooperative regulatory model contains an additional higher-order term:

\[
f_t(S_t) =
\sum_{r_i \in S_t} f_t(r_i)
+
g_t(S_t)
\]

Here, $g_t(S_t)$ captures the non-additive contribution of the regulator set. When $g_t(S_t)$ is nonzero, the regulator set carries information that is not recoverable from individual regulators alone.

This definition is consistent with biological models of transcriptional regulation in which transcription factors can act cooperatively through direct protein-protein interactions, shared cofactors, chromatin remodeling, assisted loading, motif grammar, or collective enhancer occupancy. Therefore, in this paper, cooperativity refers both to a biological property of transcriptional regulation and to an operational machine-learning target: the recovery of regulator sets whose collective activity contains information beyond pairwise interactions.

\subsection{Why Cooperative Recovery Is Difficult}

Cooperative regulator-set recovery is difficult because gene regulation is combinatorial, non-additive, context-dependent, and sparse. A target gene may depend on a small subset of regulators whose joint effect cannot be decomposed into independent regulator-target edges. As a result, individual regulators may appear weak when evaluated alone, while incorrect high-correlation regulators may dominate pairwise rankings. This creates three coupled computational challenges: combinatorial explosion in the set space, weak marginal signals for true regulators, and global competition under sparsity.

\subsection{Challenges in Higher-Order Recovery}

Three major challenges arise:

\begin{enumerate}
\item Candidate retrieval.
\item Set-level discrimination.
\item Combinatorial decoding.
\end{enumerate}

These challenges become increasingly severe as regulator-set size grows.

\section{Operational Information-Flow Interpretation of BRIDGE Bottlenecks}
\label{app:information_theoretic_bridge}

\subsection{Motivation}

Exact regulator-set recovery can fail for several distinct reasons: the true
regulators may not be retrieved, the scoring model may rank a near-miss set
above the true set, or an approximate decoder may fail to recover a high-scoring
set under combinatorial constraints. This appendix provides the detailed
mathematical definitions underlying TRACE, including retrieval loss, scoring
loss, true-set rank, decoding loss, and their operational interpretation within
BRIDGE.

We also provide an operational information-flow interpretation of these
bottlenecks using quantities that are directly measurable in our experiments.

We use the information bottleneck perspective as a conceptual framing: a useful
representation should discard irrelevant variation while preserving
task-relevant information \cite{Tishby2000InformationBottleneck}. In this
setting, the task-relevant information is the identity of the true regulator set
for a target gene. For target gene $t$, we denote the true regulator set by
$S_t$.

This appendix does not attempt to estimate high-dimensional mutual information
directly. Instead, it interprets TRACE quantities such as coverage, score gap,
true-set rank, and decoding agreement as operational measures of where
recovery-relevant information is lost, obscured, or made inaccessible.

\subsection{Search-Space Entropy}

For a target gene $t$, let $C_t$ denote the candidate regulator pool and let
$R = |S_t|$ denote the regulator-set size. If all regulator sets of size $R$
within a pool of size $M = |C_t|$ are treated as initially possible under a
uniform prior, then the uncertainty of the search problem can be written as

\[
H_R(M) = \log \binom{M}{R}.
\]

Equivalently, $H_R(M)$ is the log-size of the regulator-set search space. As \(M\) or \(R\) increases, the number of possible regulator sets grows
combinatorially, increasing the decoding burden. This motivates reporting
not only exact match, but also the number of candidate subsets evaluated and the
runtime of each decoding strategy.
\begin{table}[H]
\centering
\footnotesize
\setlength{\tabcolsep}{2.4pt}
\renewcommand{\arraystretch}{1.12}
\caption{Search-space entropy derived from the candidate-cap and runtime sensitivity experiment on SERGIO DS3, seed 42.}
\label{tab:appendix-search-space-entropy}
\begin{tabular}{@{}cccccccc@{}}
\toprule
\(R\) 
& \(M\) 
& Targets 
& \shortstack{Sets/\\target}
& \shortstack{Total\\sets}
& \shortstack{\(H_R(M)\)\\nats}
& \shortstack{\(H_R(M)\)\\bits}
& Exact \\
\midrule
2 & 20 & 57 & 190 & 10,830 & 5.247 & 7.570 & 0.263 \\
2 & 40 & 57 & 780 & 44,460 & 6.659 & 9.607 & 0.281 \\
2 & 55 & 57 & 1,485 & 84,645 & 7.303 & 10.536 & 0.281 \\
2 & 80 & 57 & 3,160 & 180,120 & 8.058 & 11.626 & 0.263 \\
\midrule
3 & 20 & 39 & 1,140 & 44,460 & 7.039 & 10.155 & 0.282 \\
3 & 40 & 39 & 9,880 & 385,320 & 9.198 & 13.270 & 0.308 \\
3 & 55 & 39 & 26,235 & 1,023,165 & 10.175 & 14.679 & 0.333 \\
3 & 80 & 39 & 82,160 & 3,204,240 & 11.316 & 16.326 & 0.333 \\
\midrule
4 & 20 & 19 & 4,845 & 92,055 & 8.486 & 12.242 & 0.053 \\
4 & 40 & 19 & 91,390 & 1,736,410 & 11.423 & 16.480 & 0.211 \\
4 & 55 & 19 & 341,055 & 6,480,045 & 12.740 & 18.380 & 0.211 \\
\bottomrule
\end{tabular}
\end{table}

Table~\ref{tab:appendix-search-space-entropy} reports the combinatorial
search-space size induced by the candidate cap \(M\) and regulator-set size
\(R\). The number of candidate sets grows rapidly with \(R\), even when \(M\)
is capped. For example, the \(R=4, M=55\) setting requires evaluating
341,055 candidate sets per target, corresponding to \(H_R(M)=12.740\) nats.
Although increasing \(M\) generally improves coverage, exact recovery does not
increase monotonically. This supports the BRIDGE interpretation that larger
candidate pools increase the decoding burden and retrieval ceiling, but do not
by themselves resolve the set-level scoring bottleneck.

\subsection{Retrieval Bottleneck}

Retrieval determines whether the true regulator set is present in the candidate
pool. For target gene $t$, retrieval success is defined as

\[
A_t =
\mathbf{1}\{S_t \subseteq C_t\}.
\]

The retrieval loss is

\[
L_{\mathrm{ret}}
=
1 -
\frac{1}{T}
\sum_{t=1}^{T} A_t.
\]

Here, $T$ is the number of evaluated target genes. This quantity is equivalent
to one minus Coverage@$M$. If $L_{\mathrm{ret}}$ is high, the true regulator set
is often absent from the candidate pool, and no downstream scoring or decoding
method can fully recover it. From an information-flow perspective, the
candidate pool has discarded information required for exact recovery.

\subsection{Scoring Bottleneck}

Scoring determines whether the model assigns a higher score to the true
regulator set than to competing incorrect sets. For each covered target gene
$t$, we define the strongest incorrect candidate set as

\[
S_t^{-}
=
\mathop{\mathrm{arg\,max}}_
{\substack{
S \subseteq C_t,\ |S|=R \\
S \neq S_t
}}
\mathrm{Score}(S,t).
\]

The score gap is defined as

\[
\mathrm{Gap}_t
=
\mathrm{Score}(S_t,t)
-
\mathrm{Score}(S_t^{-},t).
\]

A positive score gap indicates that the model ranks the true regulator set
above the strongest incorrect candidate set in the evaluated pool. A negative
score gap indicates that at least one wrong set receives a higher score than
the true set. When ties occur, we treat zero gap conservatively as an
ambiguous scoring case.

We define the conditional scoring failure indicator as

\[
B_t =
\mathbf{1}\{\mathrm{Gap}_t \leq 0\}.
\]

The conditional scoring loss among covered targets is

\[
L_{\mathrm{score}}^{\mathrm{cond}}
=
\frac{1}{|\mathcal{T}_{\mathrm{covered}}|}
\sum_{t \in \mathcal{T}_{\mathrm{covered}}} B_t,
\]

where

\[
\mathcal{T}_{\mathrm{covered}}
=
\{t : S_t \subseteq C_t\}.
\]

This isolates scoring failure from retrieval failure. In the main TRACE
decomposition, we also report the population-level scoring loss,

\[
\begin{aligned}
L_{\mathrm{score}}^{\mathrm{pop}}
&=
\mathrm{Coverage}
\times
(1-\mathrm{ConditionalExact}) \\
&=
\mathrm{Coverage}
-
\mathrm{UnconditionalExact}.
\end{aligned}
\]
Thus, conditional scoring loss measures how often scoring fails after
successful retrieval, whereas population-level scoring loss measures the total
fraction of evaluated targets lost due to scoring after accounting for
coverage.

\subsection{True-Set Rank}

The score gap gives a binary view of scoring success, but it does not show how
close the true set was to being recovered. When exhaustive scoring is feasible, we therefore also compute the rank of the
true regulator set.

For each covered target gene $t$, let $N_t$ denote the number of incorrect
candidate sets that receive a score greater than or equal to the score of the
true regulator set. The true-set rank is defined as

\[
\mathrm{Rank}_t
=
1 + N_t.
\]

If the true-set rank equals 1, the true regulator set is the highest-scoring
set in the pool. Small ranks indicate near-miss behavior, while large ranks
indicate a stronger scoring bottleneck. True-set rank therefore provides a
graded measure of how much set-level information is obscured by competing
near-miss alternatives.

\subsection{Decoding Bottleneck}

Decoding determines whether the search procedure can recover the highest-scoring
set efficiently. When exhaustive enumeration is feasible, it evaluates every
size-\(R\) subset in the capped pool and therefore removes approximate search as
a confounder. In this case, if the true set is present in the pool but not
selected, the failure is attributed to scoring rather than decoding.

Approximate decoding methods, such as beam search or proposal-based reranking,
can introduce an additional decoding bottleneck. When both exhaustive and
approximate decoded-set scores are available, let
\(\widehat{S}_t^{\mathrm{exact}}\) be the set returned by exhaustive enumeration
and let \(\widehat{S}_t^{\mathrm{approx}}\) be the set returned by an approximate
decoder. Approximate decoding failure can then be defined as

\[
D_t =
\mathbf{1}
\left\{
\mathrm{Score}(\widehat{S}_t^{\mathrm{approx}},t)
<
\mathrm{Score}(\widehat{S}_t^{\mathrm{exact}},t)
\right\}.
\]

This score-based definition avoids penalizing an approximate decoder when it
returns a different set with the same score as the exhaustive solution. The
corresponding decoding loss is

\[
L_{\mathrm{dec}}
=
\frac{1}{T}
\sum_{t=1}^{T} D_t.
\]

In the current exhaustive HOS2 experiments, decoding loss is zero by
construction because all candidate subsets in the capped pool are evaluated.
For proposal-based reranking, decoding loss is measured relative to exhaustive
HOS2 decoding because the proposal decoder evaluates only a reduced
PairS2-proposed subset space. In addition to this score-based audit, we report
exact-outcome agreement, helpful and harmful flips, and search-space reduction
as practical decoding diagnostics.

\subsection{Score-Based Decoding-Loss Audit}
\label{app:score_based_decoding_loss}

To audit the proposal-based decoder, we compute a representative seed-42
score-optimality comparison between exhaustive HOS2 decoding and PairS2 proposal
followed by HOS2 reranking. Exhaustive HOS2 decoding evaluates every size-\(R\)
subset inside the capped candidate pool and is therefore score-optimal within
that pool. Proposal reranking evaluates only a reduced PairS2-proposed subset
space, so it can return a set with a lower HOS2 score than the exhaustive HOS2
argmax.

\begin{table}[H]
\centering
\footnotesize
\setlength{\tabcolsep}{2.8pt}
\renewcommand{\arraystretch}{1.12}
\caption{Seed-42 score-based decoding-loss audit for PairS2 proposal followed by HOS2 reranking.}
\label{tab:appendix-score-based-decoding-loss}
\begin{tabular}{@{}cccccc@{}}
\toprule
\(R\)
& \shortstack{Exhaustive\\Exact}
& \shortstack{Proposal\\Exact}
& \shortstack{Decoding\\Loss}
& \shortstack{Same\\Outcome}
& \shortstack{Search\\Reduction} \\
\midrule
2 & 0.263 & 0.263 & 0.263 & 0.930 & 93.99\% \\
3 & 0.333 & 0.359 & 0.154 & 0.974 & 97.20\% \\
4 & 0.211 & 0.211 & 0.316 & 1.000 & 96.29\% \\
\bottomrule
\end{tabular}

\end{table}

All seven harmful flips occurred when proposal-stage pruning removed
at least one true regulator from the reduced candidate pool, indicating
that these failures arise from proposal coverage rather than HOS2
reranking.

\subsection{Operational Information Flow}

BRIDGE therefore interprets exact recovery through a stage-wise information-flow
lens:

\[
X
\rightarrow
C_t
\rightarrow
\mathrm{Score}(S,t)
\rightarrow
\widehat{S}_t.
\]

The measurable bottlenecks are:

\begin{itemize}
\item Retrieval loss: true regulators are missing from the candidate pool.
\item Scoring loss: the true set is present but receives a lower score than a
wrong set.
\item Decoding burden or decoding loss: approximate search may fail to recover
the exhaustive-search solution, while even exact search can become expensive as
\(\binom{M}{R}\) grows.
\end{itemize}

This operational formulation avoids requiring explicit estimation of
high-dimensional mutual information while preserving the main
information-theoretic intuition: exact recovery fails when task-relevant
information about $S_t$ is discarded during retrieval, obscured during scoring,
or made inaccessible during decoding.

\section{Cooperativity-Controlled DAG-Style Synthetic Benchmark}
\label{app:cooperativity-simulation}

A central objective of this study is to understand how recoverability changes
as biological regulation becomes increasingly cooperative. Rather than
evaluating only a small number of discrete datasets, this benchmark enables
systematic manipulation of higher-order structure through a controlled
cooperativity parameter.

This appendix provides implementation details for the cooperativity-controlled
synthetic benchmark used in Experiment~1. The benchmark is designed to isolate
the effect of cooperative, non-additive regulator-set structure on exact
regulator-set recovery. It is a controlled DAG-style synthetic GRN benchmark
rather than a SERGIO simulation. SERGIO is used as a separate synthetic GRN
benchmark, whereas this benchmark is constructed to directly vary the degree of
cooperativity while keeping the regulator-to-target structure controlled.

\subsection{DAG-Style Ground-Truth Structure}

The simulated expression matrix contains $N=320$ genes and $M=2000$ samples. In the implementation, samples are rows and genes are columns, so the matrix has size

\[
X \in \mathbb{R}^{2000 \times 320}.
\]

The first 80 genes are treated as candidate regulators,

\[
\mathcal{R} = \{G_0,\ldots,G_{79}\},
\]

and the remaining 240 genes are treated as target genes,

\[
\mathcal{T} = \{G_{80},\ldots,G_{319}\}.
\]

Each target gene $g_t \in \mathcal{T}$ is assigned a fixed parent regulator set $S_t \subset \mathcal{R}$ of size

\[
|S_t| = R = 3.
\]

This induces a directed regulator-to-target graph in which edges point from regulators to targets:

\[
r \rightarrow g_t, \quad r \in S_t.
\]

Because regulators only point to target genes and target genes do not regulate other targets in this controlled benchmark, the resulting structure is DAG-style and bipartite. Across 240 targets with three regulators per target, the benchmark contains 720 directed ground-truth regulator-target edges and 240 ground-truth cooperative regulator sets.

\subsection{Expression Generation}

For each random seed, regulator expression values are sampled from a standard normal distribution and standardized gene-wise. Parent sets and additive weights are fixed across cooperativity levels for the same seed, so that changes across $C$ reflect changes in regulatory mechanism rather than changes in graph topology.

For a target gene $g_t$ with parent regulator set

\[
S_t = \{r_1,r_2,r_3\},
\]

let $x_{r_j}$ denote the expression vector of regulator $r_j$. The additive signal is defined as

\[
a_t = \sum_{j=1}^{3} w_j x_{r_j},
\]

where $w_j$ are sampled regulator weights. The additive signal is standardized before target generation.

For cooperative targets, the corrected benchmark generates a target-specific
random nonlinear signal
\[
h_t = f_t(x_{r_1},x_{r_2},x_{r_3}),
\]
where \(f_t\) is a fixed random nonlinear MLP for target \(t\). The target
signal is
\[
y_t = \mathrm{sign}_t h_t + \lambda a_t + \epsilon_t,
\]
where \(\lambda=0.15\) and \(\epsilon_t\sim \mathcal{N}(0,0.25^2)\).

where $\mathrm{sign}_t \in \{-1,1\}$, $\lambda=0.15$ is the additive leak, and $\epsilon_t \sim \mathcal{N}(0,0.25^2)$ is Gaussian noise.

For additive targets, expression is generated as

\[
y_t = a_t + \epsilon_t.
\]

All target expression vectors are standardized after noise is added.

\subsection{Cooperativity Parameter}

The cooperativity parameter

\[
C \in \{0.0,0.2,0.4,0.6,0.8,1.0\}
\]

controls the fraction of target genes generated by the cooperative mechanism. Thus, $C=0.0$ corresponds to a fully additive benchmark, while $C=1.0$ corresponds to a fully cooperative benchmark. For each random seed, the ordering of targets assigned to the cooperative mechanism is fixed and nested across $C$. Therefore, increasing $C$ adds additional cooperative targets without changing the previously assigned parent sets or regulator expression matrix.

\subsection{Candidate Pools and Decoding}

To isolate the scoring problem, candidate pools are constructed to always contain the true regulator set together with randomly sampled distractor regulators. Each target has a candidate pool of size

\[
M = 30.
\]

Since the true regulator set has size $R=3$, each pool contains the three true regulators and 27 distractors. This gives coverage equal to one by construction:

\[
\mathrm{Coverage} = 1.0.
\]

During exhaustive decoding, all size-three subsets of the candidate pool are scored:

\[
\binom{30}{3} = 4060
\]

candidate regulator sets per target. The top-scoring set is selected as the predicted regulator set. This design controls retrieval and isolates whether the scoring model can rank
the true regulator set above incorrect candidate sets under mechanism mismatch.

\subsection{Train/Validation/Test Split}

Targets are split into train, validation, and test sets using a 60/20/20 split. The split is stratified by target type when both additive and cooperative targets are present. All reported cooperativity results use five independently generated synthetic systems with random seeds

\[
42,43,44,45,46.
\]

\subsection{Scoring Models in the Corrected Benchmark}

The corrected mechanism-mismatch benchmark compares decomposable PairS2 with
Residual HOS2 under the same oracle-covered candidate pools. PairS2 scores a
candidate regulator set by aggregating independent regulator--target evidence.
Residual HOS2 uses the same pairwise backbone but adds a learned set-level
residual correction:
\[
\mathrm{Score}(S,t)=\phi_{\mathrm{pair}}(S,t)+\psi_{\mathrm{set}}(S,t).
\]
Both models operate directly on raw expression vectors.

This differs from an earlier diagnostic proxy in which cooperative targets were
product-like and HOS2 included product-correlation features. That earlier design
could introduce feature--mechanism circularity. The corrected benchmark removes
this shortcut: cooperative targets are generated by target-specific random
nonlinear MLP mechanisms, and product-correlation features are not used. The
experiment therefore tests whether learned set-level scoring improves recovery
under mechanism mismatch.

\subsection{Implementation Parameters}
Table~\ref{tab:cooperativity-implementation-details} summarizes the fixed
implementation parameters used in the cooperativity-controlled synthetic
benchmark. These settings were held constant across all cooperativity levels so
that changes in recovery performance reflect changes in the target-generation
mechanism rather than changes in graph size, candidate-pool size, training
configuration, or decoding complexity. In particular, the candidate pool always
contains the true regulator set plus distractors, allowing the experiment to
isolate set-level scoring under a controlled retrieval setting.

\begin{table*}[t]
\centering
\caption{Implementation details for the corrected leak-free mechanism-mismatch cooperativity benchmark.}
\small
\setlength{\tabcolsep}{3pt}
\renewcommand{\arraystretch}{0.92}
\begin{tabular}{@{}p{0.55\linewidth}p{0.38\linewidth}@{}}\toprule
Parameter & Value \\
\midrule
Number of samples & 2000 \\
Number of genes & 320 \\
Number of regulators & 80 \\
Number of target genes & 240 \\
Expression matrix size & $2000 \times 320$ \\
Regulator genes & $G_0$--$G_{79}$ \\
Target genes & $G_{80}$--$G_{319}$ \\
Parent set size & $R=3$ \\
Ground-truth regulator sets & 240 \\
Ground-truth directed edges & 720 \\
Cooperativity levels & \(\{0.0,0.2,0.4,\newline 0.6,0.8,1.0\}\) \\
Noise standard deviation & 0.25 \\
Cooperative mechanism & Target-specific random nonlinear MLP \\
Product-correlation features & Not used \\
Benchmark type & Oracle-covered scoring isolation \\
Additive background coefficient for cooperative targets & 0.15\\
Candidate pool size & 30 \\
Distractors per target & 27 \\
Candidate sets per target & $\binom{30}{3}=4060$ \\
Train/validation/test split & 60/20/20 \\
Random seeds & 42--46 \\
Training epochs & 10 \\
Batch size & 128 \\
Learning rate & $10^{-3}$ \\
Weight decay & $10^{-4}$ \\
Negative sets per positive set & 8 \\
\bottomrule
\end{tabular}
\label{tab:cooperativity-implementation-details}
\end{table*}
\FloatBarrier

\section{External Edge-Ranking Baseline Results}
\label{app:external_baselines}

Table~\ref{tab:external_vs_bridge} provides an expanded TRACE-lite comparison
between external pairwise edge-ranking baselines and BRIDGE set-recovery models.
The external baselines are converted into candidate pools using their
regulator--target rankings, but they do not explicitly model regulator sets.
This comparison shows that pairwise association methods can recover some
individual regulators for \(R=2\), but their exact set recovery collapses for
larger regulator sets. In contrast, BRIDGE retrieval maintains high candidate
coverage across all \(R\), and set-level scoring with PairS2 and Residual HOS2
substantially improves exact regulator-set recovery.

This appendix is included to make the TRACE-lite comparison transparent. The
external baselines are evaluated only through their ranked regulator lists, so
their results should be interpreted as output-level recovery behavior rather
than as evidence about their internal modeling capacity. The key comparison is
therefore not whether these methods were designed for cooperative set recovery,
but whether their edge rankings are sufficient to recover complete regulator
sets under the same top-\(M\)/top-\(R\) conversion used in the main text.
\begin{table}[t]
\centering
\footnotesize
\setlength{\tabcolsep}{3.2pt}
\renewcommand{\arraystretch}{1.05}
\caption{TRACE-lite comparison of external pairwise baselines and BRIDGE/HOS2
set-recovery models on SERGIO DS3. Values are mean \(\pm\) standard deviation
across three split seeds.}
\label{tab:external_vs_bridge}
\begin{tabular}{@{}llcccc@{}}
\toprule
Class & Method & \(R\) & \(M\) & Cov.@\(M\) & Exact \\
\midrule
Pairwise & Correlation & 2 & 80 & \(0.620 \pm 0.010\) & \(0.018 \pm 0.018\) \\
Pairwise & Mutual information & 2 & 80 & \(0.345 \pm 0.027\) & \(0.023 \pm 0.010\) \\
BRIDGE retrieval & Stage1TopR & 2 & 80 & \(0.877 \pm 0.035\) & \(0.029 \pm 0.037\) \\
Set recovery & PairS2 & 2 & 80 & \(0.877 \pm 0.035\) & \(0.228 \pm 0.018\) \\
Set recovery & Residual HOS2 & 2 & 80 & \(0.877 \pm 0.035\) & \(0.287 \pm 0.073\) \\
\midrule
Pairwise & Correlation & 3 & 80 & \(0.137 \pm 0.082\) & \(0.000 \pm 0.000\) \\
Pairwise & Mutual information & 3 & 80 & \(0.043 \pm 0.015\) & \(0.000 \pm 0.000\) \\
BRIDGE retrieval & Stage1TopR & 3 & 80 & \(0.932 \pm 0.015\) & \(0.017 \pm 0.015\) \\
Set recovery & PairS2 & 3 & 80 & \(0.932 \pm 0.015\) & \(0.333 \pm 0.026\) \\
Set recovery & Residual HOS2 & 3 & 80 & \(0.932 \pm 0.015\) & \(0.350 \pm 0.030\) \\
\midrule
Pairwise & Correlation & 4 & 55 & \(0.053 \pm 0.053\) & \(0.000 \pm 0.000\) \\
Pairwise & Mutual information & 4 & 55 & \(0.000 \pm 0.000\) & \(0.000 \pm 0.000\) \\
BRIDGE retrieval & Stage1TopR & 4 & 55 & \(0.842 \pm 0.091\) & \(0.000 \pm 0.000\) \\
Set recovery & PairS2 & 4 & 55 & \(0.842 \pm 0.091\) & \(0.088 \pm 0.110\) \\
Set recovery & Residual HOS2 & 4 & 55 & \(0.842 \pm 0.091\) & \(0.211 \pm 0.000\) \\
\bottomrule
\end{tabular}%
\end{table}

\FloatBarrier
\section{Supplementary TRACE-lite Evaluation of External Foundation-Model Baselines}
\label{app:trace_lite_foundation_models}

BRIDGE is designed for recovery pipelines with explicit retrieval, set-scoring,
and decoding stages. For such models, full TRACE can be applied architecturally:
candidate-pool construction defines retrieval, the set-level scorer defines
scoring, and the search procedure defines decoding. However, many external
baselines do not expose these components directly. This includes classical
edge-ranking GRN methods, which output regulator-target edge scores, and
single-cell foundation models, which typically produce contextual gene or cell
representations rather than exact regulator sets.

To evaluate such models under the same recovery objective, we define a
TRACE-lite protocol. TRACE-lite is an operational diagnostic interface rather
than an architectural attribution method. For each target gene $t$, an external
model is first converted into a ranked list of candidate regulators using a
transparent regulator-target scoring rule. The top-$M$ regulators define an
operational candidate pool,
\[
C_t(M) = \text{top-}M \text{ regulators for target } t .
\]
Given a ground-truth or curated regulator set $S_t$, retrieval success is
defined as
\[
S_t \subseteq C_t(M).
\]
If $S_t$ is absent from $C_t(M)$, the model exhibits an operational retrieval
failure under the chosen top-$M$ threshold. If $S_t$ is present but the final
top-ranked predicted set does not match $S_t$, the failure is attributed to
operational set-ranking or selection rather than candidate absence. This
attribution should not be interpreted as evidence that the external model
contains explicit internal retrieval or set-scoring modules.

For classical GRN baselines such as GENIE3 or GRNBoost2, the model output is
already a ranked list of regulator-target edge scores. For each target gene,
the top-$M$ transcription factors are used as the candidate pool, and the top
$R$ regulators define the predicted regulator set. For single-cell foundation
models such as Geneformer, scGPT, or scFoundation, the native outputs are not
regulator sets. Therefore, model-derived embeddings, attention-derived
association scores, perturbation-derived influence scores, or gene-program
associations must first be converted into regulator-target rankings before
TRACE-lite is applied.

This conversion is intentionally treated as part of the evaluation protocol.
Attention-derived scores are interpreted only as operational association
scores, not as causal regulatory edges. Similarly, embedding similarity or
perturbation-derived scores are used to rank candidate regulators, but they do
not by themselves establish cooperative regulation. TRACE-lite therefore asks
whether the external representation supports exact regulator-set recovery under
a specified conversion rule.

Pretrained single-cell foundation models also require compatible biological
gene vocabularies. Therefore, they should not be applied to anonymized
synthetic genes such as $G_0,\ldots,G_{319}$ by artificial remapping. Instead,
foundation-model TRACE-lite evaluation should be conducted on real human
single-cell RNA-seq data with valid gene identifiers, such as gene symbols or
Ensembl IDs. When curated TF-target references are used to define $S_t$, the
resulting benchmark should be interpreted as a curated TF-set recovery proxy,
because most such references provide edge-level evidence rather than direct
experimental confirmation of cooperative regulator sets.

Under TRACE-lite, we report the same set-level recovery metrics used in the
main BRIDGE evaluation whenever applicable:
Coverage@$M$, exact set recovery, conditional exact recovery, regulator recall,
Jaccard similarity, and true-set rank. Edge-level metrics such as AUROC, AUPRC,
and Precision@$K$ may also be reported for classical GRN and foundation-model
baselines, but the primary comparison remains exact regulator-set recovery.

This supplementary analysis broadens the applicability of TRACE without
changing the main BRIDGE claims. Full TRACE is used for BRIDGE/HOS2 because
the model exposes retrieval, set-scoring, and decoding stages. TRACE-lite is
used for external baselines because their outputs must first be converted into
operational regulator-target rankings.

\begin{table}[t]
\centering
\caption{TRACE-lite interface for external GRN and foundation-model baselines.
External models are converted into regulator-target rankings before set-level
recovery metrics are computed. BRIDGE/HOS2 is included for contrast because it
exposes explicit retrieval, set-scoring, and decoding stages and is evaluated
with full TRACE.}
\label{tab:trace_lite_external_baselines}
\small
\setlength{\tabcolsep}{2.5pt}
\renewcommand{\arraystretch}{1.18}
\begin{tabular}{@{}p{0.18\linewidth}p{0.20\linewidth}p{0.27\linewidth}p{0.27\linewidth}@{}}
\toprule
Model class & Example models & Native output & TRACE / TRACE-lite conversion \\
\midrule
Classical GRN edge-ranking &
GENIE3, GRNBoost2 &
Ranked TF-target edge scores &
Top-$M$ regulators per target \\
\midrule
Simple association baseline &
Correlation, mutual information &
Pairwise gene-gene association scores &
Top-$M$ TFs by association with target \\
\midrule
Foundation-model embedding baseline &
Geneformer, scGPT, scFoundation &
Gene or cell embeddings &
Regulator-target ranking by embedding similarity \\
\midrule
Foundation-model attention baseline &
Geneformer, scGPT, when available &
Attention weights or token interactions &
Attention-derived association ranking \\
\midrule
Foundation-model perturbation baseline &
Geneformer, scGPT, scFoundation &
Predicted response to masking or perturbation &
Regulator influence score for each target \\
\midrule
BRIDGE/HOS2 &
Residual HOS2; PairS2 proposal + HOS2 rerank &
Explicit retrieval, set scoring, and decoding &
Full TRACE rather than TRACE-lite \\
\bottomrule
\end{tabular}
\end{table}

\section{Detailed Experimental Protocols}
\label{app:experimental-protocols}
\subsection{Experiment 0: Multi-Seed Baseline Re-Evaluation}
We first re-evaluate the earlier two-stage recovery pipeline across seeds
\(\{42,43,44,45,46\}\) under the BRIDGE/TRACE diagnostic protocol. This establishes the reference performance of Stage1TopR, PairS2, and Residual HOS2 before introducing BRIDGE interventions, while also auditing the stability of the earlier recovery claims.
The present study should be interpreted as a BRIDGE/TRACE re-evaluation of the
earlier two-stage recovery pipeline rather than as a direct numerical
reproduction. Although the earlier workshop version reported stronger Residual
HOS2 exact-recovery gains, the current diagnostic evaluation shows a more
size-dependent pattern: Residual HOS2 improves some settings, but exact
complete-set recovery remains limited by set-level misranking. This motivates
the shift from presenting HOS2 as a standalone recovery solution to using
BRIDGE/TRACE as a bottleneck-aware diagnostic framework.

Across five seeds, the baseline re-evaluation gives a more stability-focused view of the earlier two-stage recovery behavior. Stage1TopR achieves high candidate coverage but very low exact recovery, indicating that retrieval alone is insufficient for exact regulator-set identification. PairS2 substantially improves exact recovery over Stage1TopR, showing that learned selection within the retrieved pool is necessary. Residual HOS2 improves over PairS2 for \(R=2\) and \(R=4\), increasing unconditional exact match from \(0.218\) to \(0.298\) for \(R=2\) and from \(0.105\) to \(0.221\) for \(R=4\). For \(R=3\), PairS2 and Residual HOS2 are comparable, with PairS2 at \(0.339\) and Residual HOS2 at \(0.318\). These results show that recovery remains incomplete even when candidate coverage is high, motivating the BRIDGE bottleneck analysis. Compared with the earlier work, these results support a more cautious conclusion: Residual HOS2 can improve recovery in some settings, but exact complete-set recovery remains size-dependent and unstable, motivating the BRIDGE/TRACE bottleneck analysis.

\begin{table}[t]
\centering
\footnotesize
\setlength{\tabcolsep}{3.2pt}
\renewcommand{\arraystretch}{1.08}
\caption{Five-seed baseline re-evaluation on SERGIO DS3. Results are reported as mean \(\pm\) standard deviation across seeds \(\{42,43,44,45,46\}\).}
\label{tab:appendix-baseline-reproduction}
\begin{tabular}{@{}lcccc@{}}
\toprule
Model & \(R\) & Coverage@\(M_R\) & Uncond. Exact & Cond. Exact \\
\midrule
Stage1TopR & 2 & \(0.874\pm0.040\) & \(0.021\pm0.029\) & \(0.025\pm0.034\) \\
Stage1TopR & 3 & \(0.938\pm0.014\) & \(0.015\pm0.014\) & \(0.016\pm0.015\) \\
Stage1TopR & 4 & \(0.863\pm0.071\) & \(0.000\pm0.000\) & \(0.000\pm0.000\) \\
\midrule
PairS2 & 2 & \(0.874\pm0.040\) & \(0.218\pm0.036\) & \(0.248\pm0.034\) \\
PairS2 & 3 & \(0.938\pm0.014\) & \(0.338\pm0.021\) & \(0.361\pm0.019\) \\
PairS2 & 4 & \(0.863\pm0.071\) & \(0.105\pm0.098\) & \(0.128\pm0.125\) \\
\midrule
Residual HOS2 & 2 & \(0.874\pm0.040\) & \(0.298\pm0.066\) & \(0.342\pm0.077\) \\
Residual HOS2 & 3 & \(0.938\pm0.014\) & \(0.318\pm0.050\) & \(0.339\pm0.057\) \\
Residual HOS2 & 4 & \(0.863\pm0.071\) & \(0.221\pm0.058\) & \(0.257\pm0.066\) \\
\bottomrule
\end{tabular}
\end{table}

We evaluate Stage1TopR, PairS2, and Residual HOS2 on SERGIO DS3 using the same regulator-set sizes $R \in \{2,3,4\}$, the same capped decoding protocol $M_R=\{80,80,55\}$, and the same random seeds $\{42,43,44\}$. For each model, we report $\mathrm{Coverage}@M_R$, unconditional exact match, conditional exact match, recall, and Jaccard similarity.
\subsubsection{Paired Statistical Inference}
\label{app:sergio-pairs2-hos2-statistics}

We conducted paired statistical inference to assess whether Residual
HOS2 consistently improves recovery over PairS2 on SERGIO DS3. PairS2
and Residual HOS2 were evaluated on identical test targets within each
seed and regulator-set size. The analysis therefore used the 15 matched
seed--$R$ blocks obtained from five seeds and
$R\in\{2,3,4\}$.

For each recovery outcome, we computed
\[
\Delta =
\operatorname{Metric}_{\mathrm{Residual\ HOS2}}
-
\operatorname{Metric}_{\mathrm{PairS2}},
\]
where positive values favor Residual HOS2. Overall effects were
estimated using fixed-effects models controlling for seed and
regulator-set size. Seed-profile bootstrap intervals were obtained by
resampling each seed's complete three-$R$ profile. Holm's procedure
controlled the family-wise error rate across Jaccard similarity,
recall, and exact recovery.

\begin{table*}[t]
\centering
\caption{Paired statistical comparison of Residual HOS2 and PairS2 on SERGIO DS3. Panel A reports equal-weight mean differences across the 15 matched seed--regulator-set-size blocks using fixed-effects models controlling for seed and $R$. Seed-bootstrap intervals resample each seed's complete three-$R$ profile. Primary $p$-values are Holm-adjusted across Jaccard, recall, and exact recovery. The final two columns test whether the method difference varies across $R$, with Holm adjustment across the three outcomes. Panel B reports exploratory five-seed results separately for each $R$, with Holm adjustment across all nine outcome--$R$ comparisons. Positive values favor Residual HOS2.}
\label{tab:sergio-pairs2-hos2-inference}
\small
\textbf{Panel A: Overall seed-blocked effects}\\[1mm]
\resizebox{\textwidth}{!}{%
\begin{tabular}{lccccccc}
\toprule
Outcome & Mean gain & Blocked 95\% CI & Seed-bootstrap 95\% CI & Holm $p$ & Blocks $+/0/-$ & $F_R(2,8)$ & Holm $p_R$ \\
\midrule
Jaccard & +0.027 & [-0.011, 0.065] & [0.003, 0.055] & $0.278$ & 9/0/6 & 3.01 & $0.318$ \\
Recall & +0.015 & [-0.022, 0.052] & [-0.003, 0.039] & $0.368$ & 8/0/7 & 2.09 & $0.318$ \\
Exact recovery & +0.059 & [0.004, 0.113] & [0.014, 0.104] & $0.116$ & 8/3/4 & 2.96 & $0.318$ \\
\bottomrule
\end{tabular}%
}
\vspace{3mm}
\textbf{Panel B: Exploratory effects by regulator-set size}\\[1mm]
\resizebox{\textwidth}{!}{%
\begin{tabular}{lcccccc}
\toprule
Outcome & $R$ & Mean gain & Five-seed 95\% CI & Seed-bootstrap 95\% CI & Holm $p$ & Seeds $+/0/-$ \\
\midrule
Jaccard & 2 & +0.051 & [-0.040, 0.143] & [-0.001, 0.115] & $0.781$ & 4/0/1 \\
Jaccard & 3 & -0.030 & [-0.082, 0.023] & [-0.063, 0.006] & $0.781$ & 1/0/4 \\
Jaccard & 4 & +0.059 & [-0.019, 0.137] & [0.010, 0.108] & $0.737$ & 4/0/1 \\
\addlinespace
Recall & 2 & +0.037 & [-0.065, 0.139] & [-0.025, 0.102] & $0.781$ & 3/0/2 \\
Recall & 3 & -0.031 & [-0.069, 0.008] & [-0.053, -0.003] & $0.735$ & 1/0/4 \\
Recall & 4 & +0.039 & [-0.018, 0.097] & [0.005, 0.074] & $0.781$ & 4/0/1 \\
\addlinespace
Exact recovery & 2 & +0.081 & [-0.005, 0.166] & [0.028, 0.133] & $0.524$ & 4/1/0 \\
Exact recovery & 3 & -0.021 & [-0.103, 0.062] & [-0.072, 0.036] & $0.781$ & 1/1/3 \\
Exact recovery & 4 & +0.116 & [-0.053, 0.285] & [0.011, 0.221] & $0.781$ & 3/1/1 \\
\bottomrule
\end{tabular}%
}
\vspace{1mm}
\parbox{\textwidth}{\footnotesize \textit{Notes.} Gains are computed as Residual HOS2 minus PairS2. Blocks or seeds $+/0/-$ report positive, tied, and negative method differences. Confidence intervals are unadjusted 95\% intervals; statistical decisions use the reported Holm-adjusted $p$-values. None of the overall, $R$-heterogeneity, or exploratory $R$-specific tests remains significant after Holm correction. Exact seed-profile sign-flip tests were also nonsignificant.}
\end{table*}
Across the 15 matched seed--$R$ blocks, Residual HOS2 produced a mean
Jaccard gain of $\Delta=0.027$, with a blocked 95\% confidence interval
of $[-0.011,0.065]$. This effect was not significant after Holm
correction ($p_{\mathrm{Holm}}=.278$). The mean recall gain was
$\Delta=0.015$, with a blocked 95\% confidence interval of
$[-0.022,0.052]$ and an adjusted value of
$p_{\mathrm{Holm}}=.368$.

The mean exact-recovery gain was $\Delta=0.059$, with a blocked 95\%
confidence interval of $[0.004,0.113]$. Although the unadjusted
fixed-effects test was nominally significant ($p=.039$), the result did
not remain significant after Holm correction
($p_{\mathrm{Holm}}=.116$). Exact recovery improved in eight of the 15
seed--$R$ blocks, declined in four, and was unchanged in three.

The descriptive effects were regulator-set-size dependent. For
$R=2$, the mean gains were $+0.051$ in Jaccard, $+0.037$ in recall,
and $+0.081$ in exact recovery. For $R=3$, the corresponding effects
were negative: $-0.030$, $-0.031$, and $-0.021$. For $R=4$, the mean
gains were $+0.059$, $+0.039$, and $+0.116$. However, none of the nine
outcome--$R$ comparisons remained significant after Holm correction.

Formal omnibus tests also did not establish significant variation in
the Residual HOS2 advantage across regulator-set sizes. The
Holm-adjusted values were $.318$ for Jaccard, recall, and exact
recovery. Exact seed-profile sign-flip tests were likewise
nonsignificant. These results therefore support a cautious
interpretation: Residual HOS2 shows positive descriptive recovery gains
for $R=2$ and $R=4$, but its advantage is not stable across
regulator-set sizes or random seeds.
\begin{figure*}[t]
    \centering

    \begin{subfigure}[t]{0.48\textwidth}
        \centering
        \includegraphics[width=\linewidth]{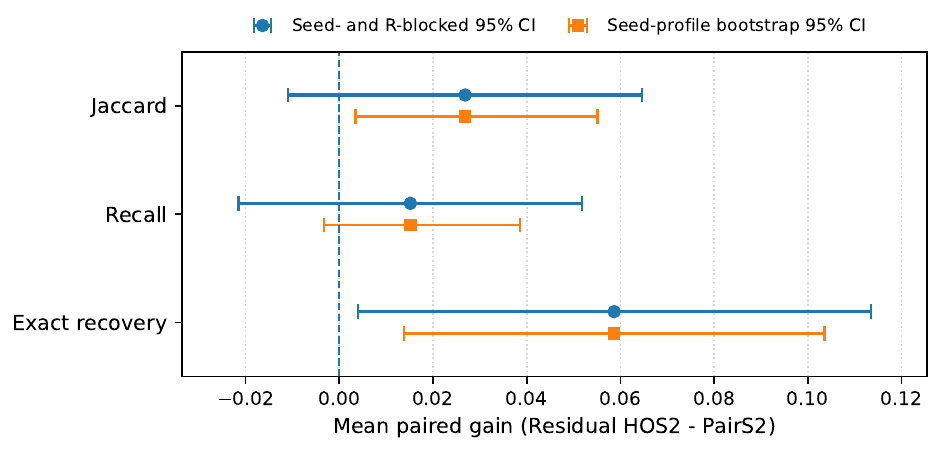}
        \caption{Overall paired differences between Residual HOS2 and PairS2 on SERGIO DS3.}
        \label{fig:sergio-pairs2-hos2-overall-inference}
    \end{subfigure}
    \hfill
    \begin{subfigure}[t]{0.48\textwidth}
        \centering
        \includegraphics[width=\linewidth]{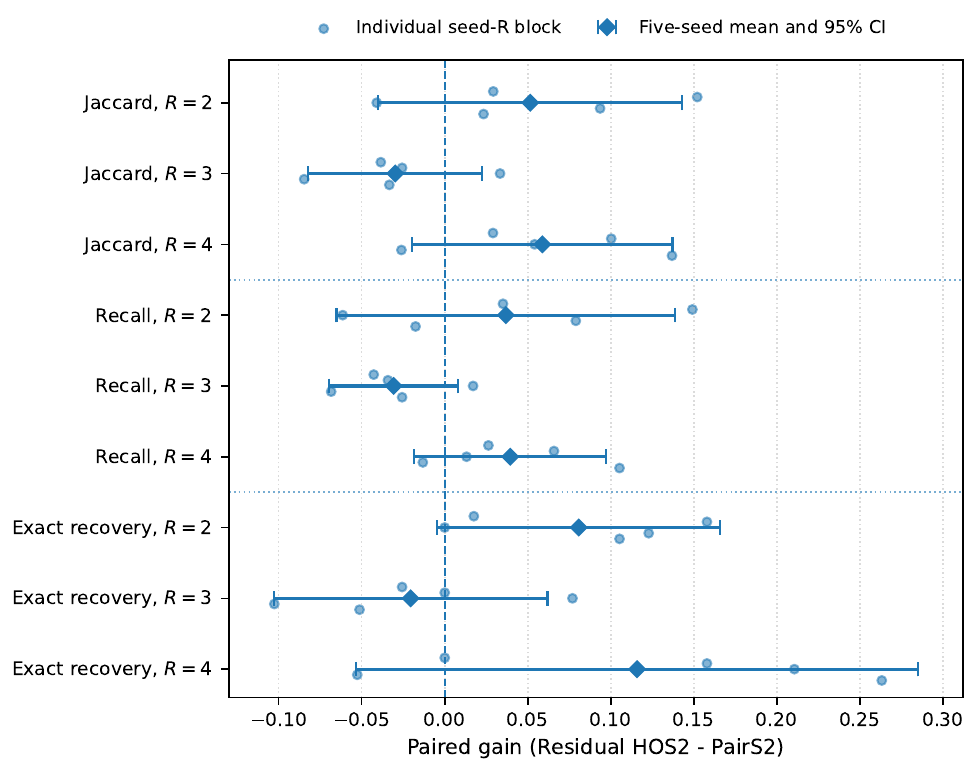}
        \caption{Regulator-set-size-specific paired differences between Residual HOS2 and PairS2 on SERGIO DS3.}
        \label{fig:sergio-pairs2-hos2-R-specific-inference}
    \end{subfigure}

    \caption{
    Paired differences between Residual HOS2 and PairS2 on SERGIO DS3.
    (a) Overall paired differences. Error bars show blocked and seed-profile
    bootstrap 95\% confidence intervals. Positive values favor Residual HOS2;
    none of the effects remains significant after Holm correction.
    (b) Regulator-set-size-specific paired differences. Small points show
    individual seed-level block differences, while diamonds and error bars show
    five-seed means and unadjusted 95\% confidence intervals. Residual HOS2
    has positive descriptive gains for $R=2$ and $R=4$ but negative gains for
    $R=3$. None of the nine size-specific tests remains significant after Holm
    correction.
    }
    \label{fig:sergio-pairs2-hos2-combined}
\end{figure*}
\FloatBarrier
\subsection{Experiment 0b: Candidate-Cap and Runtime Sensitivity}
\label{app:candidate-cap-details}

Because exact decoding depends on the capped candidate-pool size, we evaluate
recovery as a function of the decoding cap. For each regulator-set size
\(R \in \{2,3,4\}\), we vary the cap \(M\) and measure
\(\mathrm{Coverage}@M\), unconditional exact match, conditional exact match,
runtime, and the number of candidate subsets evaluated.

The central question is whether exact recovery is limited by model scoring
quality or by the computational restriction imposed by capped enumeration.

\subsubsection{Full Candidate-Cap Results}

Table~\ref{tab:appendix-cap-full} provides the full candidate-cap diagnostic
results for seed 42, including set-overlap metrics that were omitted from the
main text for space. The main paper reports Coverage@\(M\), exact recovery,
runtime, and the number of candidate sets evaluated. Here, we additionally
report mean Jaccard, recall, and precision across test targets. These values
show that larger candidate pools can improve retrieval coverage and partial
overlap, but exact recovery may still plateau. This supports the BRIDGE
interpretation that retrieval, scoring, and decoding should be diagnosed
separately.

\subsubsection{Interpretation}

The full diagnostic results show that candidate-pool expansion increases the
retrieval ceiling but does not guarantee exact recovery. For \(R=3\), increasing
\(M\) from 55 to 80 increases Coverage@\(M\) from 0.821 to 0.949, while exact
recovery remains unchanged at 0.333. For \(R=4\), increasing \(M\) from 40 to 55
increases Coverage@\(M\) from 0.684 to 0.737, while exact recovery remains
unchanged at 0.211. These plateau patterns indicate that once additional true
regulators enter the candidate pool, remaining failures may be caused by
scoring ambiguity rather than retrieval alone.

The runtime results also quantify the decoding bottleneck. For \(R=4\),
increasing \(M\) from 40 to 55 increases the number of evaluated candidate sets
from 1.74 million to 6.48 million and increases runtime from 103.5 seconds to
432.9 seconds. This confirms that exact exhaustive decoding becomes increasingly
expensive as both \(R\) and \(M\) grow, motivating the need for
bottleneck-aware diagnosis and scalable decoding strategies.

\begin{table*}[t]
\centering
\small

\caption{Full candidate-cap diagnostic results for exact HOS2 decoding on SERGIO DS3 test targets, seed 42. All runtimes were measured on CPU during exhaustive exact decoding and should be interpreted as diagnostic wall-clock measurements rather than optimized inference benchmarks.}
\label{tab:appendix-cap-full}
\begin{tabular}{ccccccccc}
\toprule
$R$ & $M$ & Coverage@$M$ & Exact & Jaccard & Recall & Precision & Runtime (s) & Candidate Sets \\
\midrule
2 & 20 & 0.526 & 0.263 & 0.374 & 0.430 & 0.430 & 0.5 & 10,830 \\
2 & 40 & 0.772 & 0.281 & 0.398 & 0.456 & 0.456 & 1.4 & 44,460 \\
2 & 55 & 0.789 & 0.281 & 0.398 & 0.456 & 0.456 & 2.5 & 84,645 \\
2 & 80 & 0.877 & 0.263 & 0.392 & 0.456 & 0.456 & 5.3 & 180,120 \\
\midrule
3 & 20 & 0.641 & 0.282 & 0.508 & 0.598 & 0.598 & 1.8 & 44,460 \\
3 & 40 & 0.769 & 0.308 & 0.533 & 0.624 & 0.624 & 18.5 & 385,320 \\
3 & 55 & 0.821 & 0.333 & 0.554 & 0.641 & 0.641 & 47.2 & 1,023,165 \\
3 & 80 & 0.949 & 0.333 & 0.549 & 0.632 & 0.632 & 167.4 & 3,204,240 \\
\midrule
4 & 20 & 0.368 & 0.053 & 0.324 & 0.434 & 0.434 & 6.7 & 92,055 \\
4 & 40 & 0.684 & 0.211 & 0.440 & 0.526 & 0.526 & 103.5 & 1,736,410 \\
4 & 55 & 0.737 & 0.211 & 0.440 & 0.526 & 0.526 & 432.9 & 6,480,045 \\
\bottomrule
\end{tabular}
\end{table*}

\begin{figure*}[t]
\centering
\includegraphics[width=\textwidth]{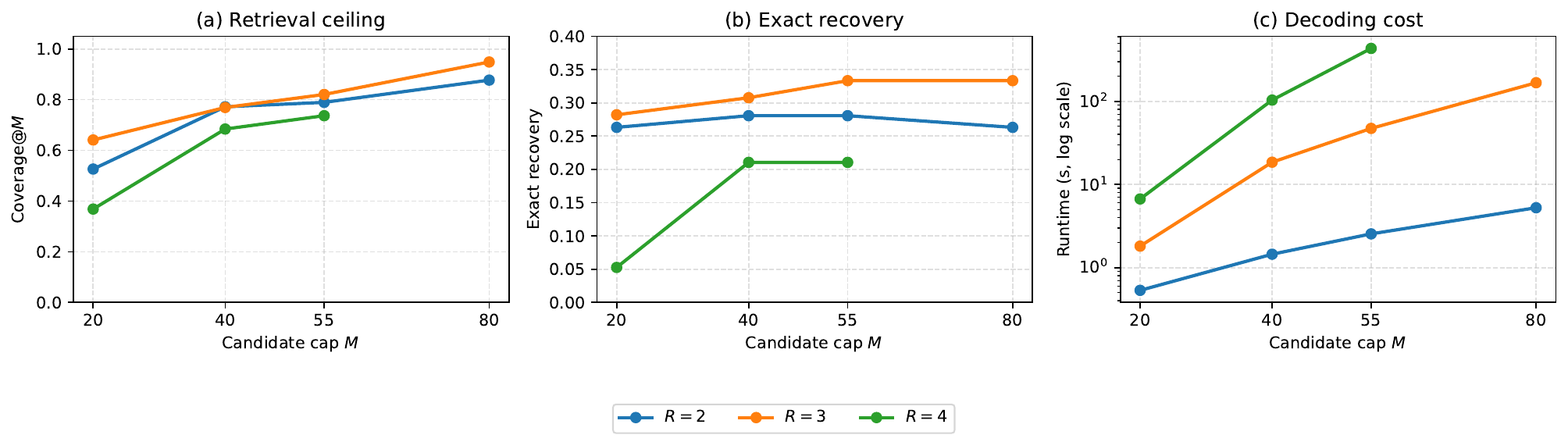}
\caption{Candidate-cap diagnostic trends for exact HOS2 decoding on SERGIO DS3 test targets, seed 42. Increasing the candidate cap \(M\) improves Coverage@\(M\), but unconditional exact recovery plateaus for larger caps while runtime grows rapidly on a log scale. This supports the BRIDGE interpretation that retrieval coverage, scoring quality, and decoding cost should be diagnosed as separate bottlenecks.}
\label{fig:candidate-cap-diagnostic}
\end{figure*}

Figure~\ref{fig:candidate-cap-diagnostic} visualizes the same candidate-cap
tradeoff. Increasing \(M\) raises the retrieval ceiling because the complete
true regulator set is more likely to enter the candidate pool, but this does not
translate monotonically into exact recovery. For \(R=3\), increasing \(M\) from
55 to 80 raises Coverage@\(M\) from 0.821 to 0.949, while exact recovery remains
fixed at 0.333. Similarly, for \(R=4\), increasing \(M\) from 40 to 55 raises
Coverage@\(M\) from 0.684 to 0.737, while exact recovery remains fixed at
0.211. At the same time, runtime grows sharply because the number of enumerated
candidate sets increases combinatorially with \(M\) and \(R\). The \(R=4, M=55\) setting evaluates 6.48 million candidate sets and requires
432.9 seconds on CPU. These results show why BRIDGE separates retrieval,
scoring, and decoding: expanding the candidate pool can improve coverage, but
once the true set is present, remaining failures are better explained by
set-level scoring ambiguity and decoding cost rather than retrieval alone.
\FloatBarrier
Table~\ref{tab:appendix-cap-full} shows that increasing $M$ improves $\mathrm{Coverage}@M$, since the true regulator set is more likely to be included in the candidate pool. However, exact recovery does not improve monotonically. For $R=3$, increasing $M$ from 55 to 80 raises coverage from 0.821 to 0.949, but exact recovery remains fixed at 0.333. Similarly, for $R=4$, increasing $M$ from 40 to 55 raises coverage from 0.684 to 0.737, while exact recovery remains fixed at 0.211. In contrast, runtime grows sharply: the $R=4, M=55$ setting evaluates 6.48 million candidate sets and requires 432.9 seconds on CPU. These results show that expanding the candidate pool improves the retrieval ceiling but does not by itself solve exact regulator-set recovery, motivating BRIDGE's separation of retrieval, scoring, and decoding bottlenecks. A derived search-space entropy analysis of these candidate caps is provided in Appendix~\ref{app:information_theoretic_bridge}, where \(H_R(M)=\log\binom{M}{R}\) is used as a log-scale measure of the combinatorial decoding space.

\subsection{Experiment 0c: Oracle Retrieval and Scoring Ceiling}
\label{app:oracle-retrieval-ceiling}
To isolate scoring from retrieval, we evaluate Residual HOS2 under an
oracle retrieval setting. For each target gene, we construct a candidate
pool by starting from the Stage-1 candidate list and injecting any
missing true regulators, followed by true-safe trimming that preserves
the complete ground-truth regulator set.

This guarantees that the true regulator set is present in the candidate
pool for every evaluated target. Residual HOS2 decoding is performed by
exact subset enumeration within the oracle-corrected pool. Therefore,
any remaining exact-recovery failure reflects set-level scoring rather
than retrieval absence or approximate decoding.

This experiment directly tests whether residual set attention improves recovery
when retrieval is no longer the limiting factor, and whether remaining failures
are due to near-miss confusers or insufficient score separation.

\subsubsection{Full Oracle Retrieval Results}

Table~\ref{tab:oracle-retrieval-scoring-ceiling-five-seed} provides the five-seed oracle-retrieval diagnostic results. The main text reports the compact comparison between normal and oracle exact recovery. Here, we additionally report baseline coverage, oracle coverage, baseline exact recovery, oracle exact recovery, and oracle Jaccard. Oracle retrieval forces the true regulator set to be present in every candidate pool, so remaining exact-recovery failures reflect set-level scoring ambiguity under exhaustive decoding rather than candidate absence.

\begin{table*}[t]
\centering
\small
\setlength{\tabcolsep}{5pt}
\renewcommand{\arraystretch}{1.15}
\caption{Five-seed oracle retrieval and scoring-ceiling results on SERGIO DS3. Results are reported as mean \(\pm\) standard deviation across seeds \(\{42,43,44,45,46\}\). Oracle coverage is \(1.0\) by construction.}
\label{tab:oracle-retrieval-scoring-ceiling-five-seed}
\begin{tabular}{@{}ccccccc@{}}
\toprule
\(R\) & \(M\) & Base Cov. & Oracle Cov. & Base Exact & Oracle Exact & Oracle Jaccard \\
\midrule
2 & 80 & \(0.874\pm0.040\) & \(1.000\pm0.000\) & \(0.298\pm0.066\) & \(0.302\pm0.064\) & \(0.416\pm0.078\) \\
3 & 80 & \(0.938\pm0.014\) & \(1.000\pm0.000\) & \(0.318\pm0.050\) & \(0.318\pm0.050\) & \(0.563\pm0.045\) \\
4 & 55 & \(0.863\pm0.071\) & \(1.000\pm0.000\) & \(0.221\pm0.058\) & \(0.221\pm0.058\) & \(0.459\pm0.047\) \\
\bottomrule
\end{tabular}
\end{table*}
Across five seeds, oracle retrieval raises Coverage@\(M\) to \(1.0\) for all regulator-set sizes, but exact recovery changes only minimally. For \(R=2\), baseline exact recovery is \(0.298\pm0.066\), while oracle retrieval gives \(0.302\pm0.064\). For \(R=3\), exact recovery remains \(0.318\pm0.050\). For \(R=4\), exact recovery remains \(0.221\pm0.058\). Thus, even when the complete true regulator set is forced into every candidate pool, exact recovery remains limited. This confirms that the dominant remaining bottleneck is set-level scoring rather than retrieval absence.
This result is central to the BRIDGE/TRACE interpretation. If exact recovery
were primarily limited by candidate absence, forcing the complete regulator set
into every candidate pool would be expected to produce a substantial increase
in exact recovery. Instead, oracle retrieval mainly removes the retrieval
ceiling while leaving the final prediction behavior nearly unchanged. The true
set is available to the decoder, but the scorer often assigns a higher score to
a near-miss alternative. This separates two failure modes that are usually
collapsed in edge-ranking evaluations: failing to retrieve the necessary
regulators and failing to select the correct regulator set once those regulators
are present. The oracle experiment therefore supports the use of score gap,
true-set rank, and conditional exact recovery as downstream diagnostics. It also
explains why simply expanding the candidate pool is insufficient: larger pools
can improve coverage, but they also introduce more competing subsets and do not
guarantee that the true set will be ranked first.

The effect is especially important for interpreting the larger regulator-set
sizes. For \(R=3\) and \(R=4\), oracle retrieval raises Coverage@\(M\) to one by
construction, but exact recovery remains unchanged relative to the baseline
setting. Thus, the limiting factor is not whether the true regulators are
available somewhere in the candidate pool, but whether the set scorer can rank
the exact combination above many plausible alternatives. This is precisely the
case where decomposable edge-level evidence can be misleading: several
regulators may be individually plausible, and several near-miss sets may share
substantial overlap with the truth, but exact recovery requires selecting the
entire regulator set jointly. In this sense, oracle retrieval converts the
experiment into a controlled scoring audit. It shows that complete regulator-set
recovery requires more than high recall; it requires reliable set-level
discrimination among overlapping candidate subsets.
More broadly, the oracle-retrieval result clarifies the role of Stage~1 in
BRIDGE. Candidate retrieval remains necessary because no scorer can recover a
true regulator set that is absent from the pool. However, once retrieval is
sufficiently strong, further improvements in coverage have diminishing returns
unless the scoring function can distinguish the true set from high-overlap
confusers. This explains why BRIDGE treats retrieval and scoring as separate
bottlenecks rather than combining them into a single end-to-end exact-recovery
number. It also motivates the later intervention analysis: if the true set is
already present but misranked, the appropriate repair is not simply to enlarge
the pool, but to diagnose score gaps, inspect true-set ranks, and test whether
scoring or decoding interventions safely improve held-out recovery.
Finally, the oracle setting also prevents an overly optimistic interpretation of
high candidate coverage. In a standard retrieval-only analysis, high
Coverage@\(M\) might suggest that the recovery problem is nearly solved because
the correct regulators are present somewhere in the candidate list. The oracle
experiment shows why this is not enough. Exact regulator-set recovery is a
ranking problem over sets, not only a recall problem over individual regulators.
When many candidate subsets share one or more true regulators, the scorer must
resolve fine-grained differences among highly similar alternatives. Therefore,
oracle coverage should be interpreted as a ceiling on what retrieval can provide,
not as evidence that the downstream recovery problem has become easy.
\FloatBarrier
Figure~\ref{fig:oracle-retrieval-comparison} shows that oracle retrieval raises Coverage@\(M\) to \(1.0\) for all regulator-set sizes, but exact recovery changes only minimally. Exact recovery changes from \(0.298\pm0.066\) to \(0.302\pm0.064\) for \(R=2\), remains \(0.318\pm0.050\) for \(R=3\), and remains \(0.221\pm0.058\) for \(R=4\). Thus, the remaining failures reflect set-level scoring ambiguity rather than candidate absence.
\FloatBarrier
\subsubsection{Paired Oracle-Retrieval Inference}
\label{app:oracle-retrieval-inference}

We paired standard and oracle-retrieval HOS2 predictions for all 575
test targets across five seeds and $R\in\{2,3,4\}$. Oracle retrieval
increased candidate coverage to one for every target, but produced only
one additional exact recovery and no exact-recovery regressions.

Across the 15 matched seed--$R$ blocks, the mean oracle-retrieval gains
were $0.0017$ in Jaccard similarity, $0.0017$ in recall, and $0.0012$
in exact recovery. None remained significant after Holm correction
($p_{\mathrm{Holm}}=.714$ for all three outcomes). The oracle effect
also did not vary significantly across regulator-set sizes. Thus,
eliminating candidate absence produced negligible recovery improvement,
supporting the TRACE attribution of the remaining failures primarily
to set-level scoring rather than retrieval.

As shown in Table~\ref{tab:oracle-retrieval-inference}, paired
inference across the 15 seed--$R$ blocks found negligible
oracle-retrieval gains: $0.0017$ in Jaccard, $0.0017$ in recall,
and $0.0012$ in exact recovery. None was significant after Holm
correction ($p_{\mathrm{Holm}}=.714$ for all three outcomes).
Across 575 paired targets, oracle retrieval produced only one
additional exact recovery and no regressions. Thus, eliminating
candidate absence did not materially improve recovery, supporting
set-level scoring as the dominant remaining bottleneck.

\subsection{Experiment 1: Recoverability versus Cooperativity}
\label{app:cooperativity-results}

This experiment evaluates the corrected leak-free mechanism-mismatch
cooperativity stress test. The goal is to measure whether learned set-level
scoring improves regulator-set recovery as the fraction of cooperatively
generated targets varies. We construct a controlled synthetic GRN benchmark in
which each target gene is assigned a regulator set of fixed size \(R=3\).
Additive targets are generated from a linear combination of their regulators,
whereas cooperative targets are generated using target-specific random nonlinear
MLP mechanisms rather than product interactions.

Datasets are generated across the cooperativity continuum:
\[
C \in \{0.0,0.2,0.4,0.6,0.8,1.0\}.
\]
The cooperativity parameter \(C\) controls the fraction of targets generated by
cooperative rather than additive mechanisms. Thus, \(C=0.0\) corresponds to a
fully additive setting, while \(C=1.0\) corresponds to a fully cooperative
setting.

To isolate set-level scoring, candidate pools are oracle-covered: each pool
contains the true regulator set together with randomly sampled distractor
regulators. This controls retrieval and allows us to evaluate whether the
scoring model ranks the true regulator set above incorrect candidate sets. We
compare PairS2 against Residual HOS2 using exact recovery, recall, Jaccard
similarity, coverage, score gap, and true-set rank. Results are reported over
five independently generated synthetic systems using random seeds \(42\)--\(46\).

\FloatBarrier
\begin{table*}[t]
\centering
\small
\setlength{\tabcolsep}{3pt}
\renewcommand{\arraystretch}{1.10}
\caption{
Corrected leak-free mechanism-mismatch cooperativity results by cooperativity
level \(C\). Values are averaged over five random seeds. Candidate pools are
oracle-covered, so the experiment isolates set-level scoring. Gains are computed
as Residual HOS2 minus PairS2.
}
\label{tab:cooperativity-recovery-by-c}
\begin{tabular}{c|ccc|ccc|ccc}
\hline
\(C\) &
Pair Exact & HOS2 Exact & \(\Delta\) Exact &
Pair Jaccard & HOS2 Jaccard & \(\Delta\) Jaccard &
Pair Recall & HOS2 Recall & \(\Delta\) Recall \\
\hline
0.0 & 0.050 & 0.112 & +0.062 & 0.388 & 0.455 & +0.068 & 0.531 & 0.592 & +0.061 \\
0.2 & 0.058 & 0.092 & +0.033 & 0.396 & 0.455 & +0.059 & 0.537 & 0.596 & +0.058 \\
0.4 & 0.054 & 0.092 & +0.037 & 0.400 & 0.447 & +0.047 & 0.543 & 0.588 & +0.044 \\
0.6 & 0.067 & 0.150 & +0.083 & 0.391 & 0.481 & +0.090 & 0.529 & 0.613 & +0.083 \\
0.8 & 0.021 & 0.125 & +0.104 & 0.329 & 0.458 & +0.128 & 0.469 & 0.592 & +0.122 \\
1.0 & 0.071 & 0.108 & +0.037 & 0.387 & 0.462 & +0.075 & 0.524 & 0.600 & +0.076 \\
\hline
\end{tabular}
\end{table*}
\FloatBarrier

Table~\ref{tab:cooperativity-recovery-by-c} summarizes the exact-recovery results
across the cooperativity continuum.
Across all cooperativity levels, Residual HOS2 improves Jaccard and recall over
PairS2. However, the gains are not monotonic in \(C\). This supports the
interpretation that learned set-level scoring improves graded regulator-set
recovery under mechanism mismatch, while exact complete-set recovery remains
difficult.

These results support a more cautious interpretation. Under the corrected
mechanism-mismatch benchmark, Residual HOS2 improves graded recovery over
PairS2 across cooperativity levels, but the gain is not monotonic in \(C\).
Exact recovery also remains low overall. Thus, learned set-level scoring
improves regulator-set overlap under mechanism mismatch, but complete exact
recovery remains a bottleneck.

\subsubsection{Statistical Inference}
\label{app:controlled-cooperativity-statistics}

We conducted paired statistical inference to quantify the improvement
of Residual HOS2 over PairS2 in the corrected leak-free
mechanism-mismatch benchmark. Each seed--cooperativity combination was
treated as a matched experimental block, giving 30 blocks from five
independently generated systems and six cooperativity levels. Within
each block, PairS2 and Residual HOS2 were evaluated on the same 48 test
targets.

For each outcome, we computed the paired difference
\[
\Delta =
\operatorname{Metric}_{\mathrm{Residual\ HOS2}}
-
\operatorname{Metric}_{\mathrm{PairS2}}.
\]
We analyzed these block-level differences using fixed-effects models
that controlled for seed and cooperativity level. We additionally
constructed 95\% cluster-bootstrap confidence intervals by resampling
each seed's complete six-level cooperativity profile. Holm's step-down
procedure controlled the family-wise error rate across the three
primary outcomes: Jaccard similarity, recall, and exact recovery.

\begin{table*}[t]
\centering
\caption{Statistical inference for Residual HOS2 relative to PairS2 in the corrected leak-free mechanism-mismatch cooperativity benchmark. Mean differences are computed over 30 matched seed--cooperativity blocks. Blocked confidence intervals are obtained from fixed-effects models controlling for seed and cooperativity level. Bootstrap intervals resample each seed's complete six-level profile. Primary $p$-values are Holm-adjusted across Jaccard, recall, and exact recovery. The final two columns report a separate Holm-adjusted family of omnibus tests evaluating whether the advantage differs across cooperativity levels. Positive values favor Residual HOS2.}
\label{tab:controlled-cooperativity-inference}
\small
\resizebox{\textwidth}{!}{%
\begin{tabular}{lccccccc}
\toprule
Outcome & Mean gain & Blocked 95\% CI & Seed-bootstrap 95\% CI & Holm $p$ & Blocks $+/0/-$ & $F_C(5,20)$ & Holm $p_C$ \\
\midrule
Jaccard & +0.078 & [0.063, 0.093] & [0.067, 0.089] & $2.36\times 10^{-9}$ & 30/0/0 & 2.57 & $0.165$ \\
Recall & +0.074 & [0.060, 0.089] & [0.061, 0.087] & $2.36\times 10^{-9}$ & 30/0/0 & 2.63 & $0.165$ \\
Exact recovery & +0.060 & [0.041, 0.079] & [0.051, 0.069] & $2.04\times 10^{-6}$ & 25/3/2 & 1.71 & $0.178$ \\
\bottomrule
\end{tabular}%
}
\vspace{1mm}\parbox{\textwidth}{\footnotesize \textit{Notes.} Blocks $+/0/-$ report the numbers of seed--cooperativity blocks in which Residual HOS2 improved, tied, or performed worse than PairS2. The exact two-sided seed-level sign-flip sensitivity test produced $p=0.0625$ for each outcome. With five seeds, this is the smallest attainable two-sided sign-flip $p$-value.}
\end{table*}

Residual HOS2 significantly improved all three primary recovery
outcomes after Holm correction. The mean Jaccard improvement was
$\Delta=0.078$ with a seed- and cooperativity-blocked 95\% confidence
interval of $[0.063,0.093]$ and a seed-cluster bootstrap interval of
$[0.067,0.089]$. The corresponding Holm-adjusted value was
$p=2.36\times10^{-9}$. Jaccard improved in all 30 matched
seed--cooperativity blocks.

Recall improved by $\Delta=0.074$, with a blocked 95\% confidence
interval of $[0.060,0.089]$ and a seed-cluster bootstrap interval of
$[0.061,0.087]$. The Holm-adjusted value was
$p=2.36\times10^{-9}$, and recall also improved in all 30 matched
blocks.

Exact recovery improved by $\Delta=0.060$, with a blocked 95\%
confidence interval of $[0.041,0.079]$ and a seed-cluster bootstrap
interval of $[0.051,0.069]$. The Holm-adjusted value was
$p=2.04\times10^{-6}$. Exact recovery improved in 25 blocks, tied in
three blocks, and declined in two blocks.

The paired target-level exact-recovery transitions provide an
additional interpretation of this effect. Across 1,440 paired target
evaluations, Residual HOS2 converted 131 PairS2 failures into exact
recoveries while losing 45 exact recoveries achieved by PairS2. This
corresponds to 86 net exact-recovery improvements and a descriptive
improvement-to-regression ratio of $131/45=2.91$.

We separately tested whether the magnitude of the Residual HOS2
advantage differed across cooperativity levels. These omnibus effects
were not significant after Holm correction: Jaccard,
$F(5,20)=2.57$, adjusted $p=.165$; recall,
$F(5,20)=2.63$, adjusted $p=.165$; and exact recovery,
$F(5,20)=1.71$, adjusted $p=.178$. Thus, Residual HOS2 showed a
consistent positive overall advantage, but the magnitude of that
advantage was not monotonic in cooperativity level $C$.

Because only five independent seeds were available, we also report an
exact seed-level sign-flip sensitivity analysis. With five seeds,
there are only $2^5=32$ possible sign assignments, so the smallest
attainable two-sided value is $p=.0625$. This minimum value was
obtained for all three outcomes. We therefore interpret the sign-flip
test as a conservative small-sample sensitivity analysis and base the
primary inference on the seed- and cooperativity-blocked models,
cluster-bootstrap confidence intervals, effect sizes, and consistency
across matched blocks.

\begin{figure*}[t]
    \centering

    \begin{subfigure}[t]{0.46\textwidth}
        \centering
        \vspace{0pt}
        \includegraphics[
            width=\linewidth
        ]{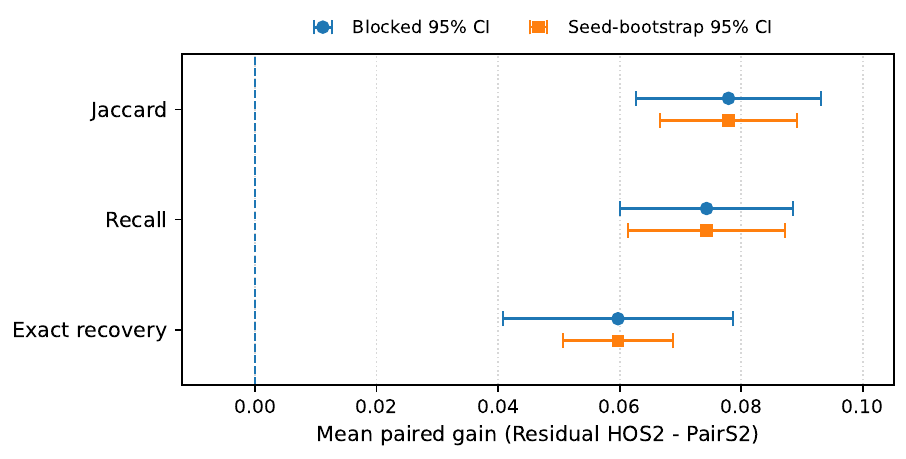}
        \caption{
        Overall paired gains of Residual HOS2 over PairS2. Circles show
        estimates and 95\% confidence intervals from models blocking
        on seed and cooperativity level; squares show seed-cluster
        bootstrap 95\% confidence intervals.
        }
        \label{fig:controlled-cooperativity-forest}
    \end{subfigure}
    \hfill
    \begin{subfigure}[t]{0.51\textwidth}
        \centering
        \vspace{0pt}
        \includegraphics[
            width=\linewidth
        ]{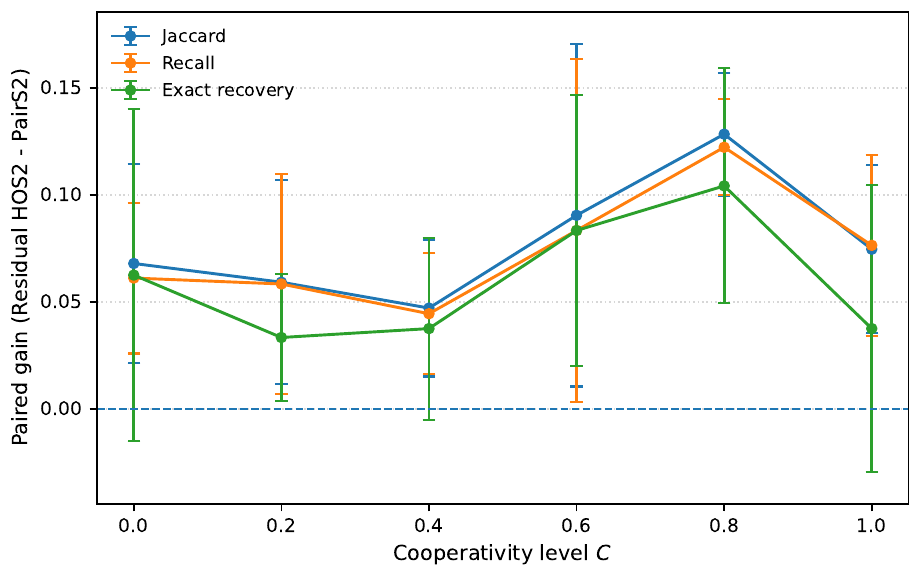}
        \caption{
        Mean paired gain at each cooperativity level. Error bars are
        descriptive 95\% $t$ intervals across the five seeds.
        }
        \label{fig:controlled-cooperativity-by-c}
    \end{subfigure}

    \caption{
    Statistical evaluation of Residual HOS2 relative to PairS2 in the
    corrected leak-free mechanism-mismatch benchmark. Positive values
    favor Residual HOS2. Overall Jaccard, recall, and exact-recovery
    gains remain positive after seed blocking and seed-cluster
    bootstrap analysis. The level-specific results show that the
    magnitude of the advantage is not monotonic in cooperativity level
    $C$.
    }
    \label{fig:controlled-cooperativity-statistics}
\end{figure*}

\FloatBarrier

\subsection{Experiment 2: Hard-Negative Intervention Study}
\label{app:hard-negative-intervention}

This experiment evaluates whether increasingly difficult negative samples improve discrimination of cooperative regulator sets.

Negative types include:

\begin{itemize}
\item Random negatives.
\item One-swap negatives.
\item Near-miss negatives.
\item Confuser negatives.
\end{itemize}

We also evaluate curriculum-based training, where progressively harder negatives are introduced during optimization.

The central question is whether hard negatives improve higher-order discrimination and exact recovery.

Because TRACE identifies scoring as the dominant bottleneck, we next evaluate whether increasing near-miss hard-negative pressure improves exact regulator-set recovery. The default HOS2 training procedure already uses a curriculum that transitions from mostly random negatives to more near-miss confuser negatives. Therefore, this experiment does not introduce hard negatives for the first time; instead, it tests whether a stronger hard-negative curriculum improves recovery.

The default curriculum uses 80\% random negatives and 20\% near-miss negatives during warm-up, followed by 20\% random negatives and 80\% near-miss negatives during later epochs. The hard-enhanced curriculum uses 20\% random negatives and 80\% near-miss negatives during warm-up, followed by 100\% near-miss negatives during later epochs.

Figure~\ref{fig:hard-negative-ablation} shows that increasing near-miss
hard-negative pressure does not produce a consistent improvement between
regulator-set sizes. For $R=2$, hard-negative enhancement increased the three-seed mean
unconditional exact recovery from $0.2865$ to $0.3216$, corresponding
to a relative increase of $12.24\%$, and increased conditional exact
recovery from $0.3292$ to $0.3660$. However, these descriptive gains
were inconsistent across seeds, and the paired unconditional
exact-recovery effect was not statistically significant. Moreover, this improvement does not generalize to larger
regulator sets. For \(R=3\), unconditional exact recovery decreases from 0.3504
to 0.2906, and for \(R=4\), it decreases from 0.2105 to 0.1754. Conditional
exact recovery follows the same pattern, decreasing for both \(R=3\) and
\(R=4\). These results indicate that hard-negative scoring repair is
size-dependent and must be validated before being accepted as a BRIDGE
intervention.

These results suggest that the TRACE-identified scoring bottleneck is not solved simply by increasing the proportion of near-miss negatives. While stronger near-miss pressure may help smaller regulator sets, it does not generalize uniformly to larger combinatorial regulator sets. Instead, hard-negative intervention must be treated as a size-dependent and validation-dependent repair. This motivates the BRIDGE intervention-selection view: a detected scoring bottleneck does not automatically imply that stronger hard-negative training should be applied. Rather, candidate scoring interventions must be tested on held-out recovery and rejected when they degrade exact regulator-set recovery.

\begin{table}[t]
\centering
\caption{Exploratory paired inference for hard-negative-enhanced HOS2 relative to baseline HOS2 on SERGIO DS3. Positive values favor the hard-enhanced scorer.}
\label{tab:hard-negative-inference}
\small
\setlength{\tabcolsep}{4.5pt}
\renewcommand{\arraystretch}{1.12}
\begin{tabular}{lccc}
\toprule
Outcome & Mean gain & Blocked 95\% CI & Holm $p$ \\
\midrule
Jaccard & $-0.015$ & $[-0.102,\ 0.072]$ & $1.000$ \\
Recall & $-0.007$ & $[-0.087,\ 0.073]$ & $1.000$ \\
Exact recovery & $-0.020$ & $[-0.137,\ 0.097]$ & $1.000$ \\
\bottomrule
\end{tabular}
\vspace{1mm}
\parbox{\columnwidth}{
\footnotesize
\textit{Notes.} Effects were estimated from nine matched seed--$R$ blocks using seeds 42--44. Confidence intervals are unadjusted; $p$-values are Holm-adjusted across Jaccard, recall, and exact recovery. All tests are exploratory because only three independent seeds were available. Exact seed-profile sign-flip tests were also nonsignificant.
}
\end{table}

As shown in Table~\ref{tab:hard-negative-inference}, exploratory paired
inference did not establish a reliable benefit from stronger
hard-negative training. Across the nine matched seed--$R$ blocks, the
mean changes were $-0.015$ in Jaccard, $-0.007$ in recall, and
$-0.020$ in exact recovery; none was significant after Holm correction.
Descriptively, the three-seed mean improved for $R=2$ but declined for
$R=3$ and $R=4$. None of the nine size-specific comparisons remained
significant after Holm correction. Across 345 paired targets, the
hard-negative-enhanced scorer produced 29 exact-recovery improvements
and 32 regressions. These results indicate that stronger hard-negative
pressure is not a uniformly reliable scoring repair and should remain
subject to held-out BRIDGE validation.

\subsection{Experiment 3: Decoder Scalability and Intervention Study}
\label{app:decoder-scalability}
This experiment investigates whether decoding bottlenecks limit exact regulator-set recovery. Exhaustive enumeration is treated as the scoring gold standard within capped pools for $R \leq 4$, because it evaluates all candidate subsets and therefore removes approximate search as a confounder. However, exhaustive decoding scales combinatorially with $M$ and $R$, making it impractical for larger regulator teams or larger candidate pools.

We compare four decoding strategies:

\begin{itemize}
\item Exhaustive enumeration over all size-$R$ subsets.
\item Beam search over incrementally constructed regulator sets.
\item PairS2 proposal followed by Residual HOS2 reranking.
\item Local swap refinement initialized from the PairS2 top-$R$ set.
\end{itemize}

For each decoder, we report exact match, recall, Jaccard similarity, runtime, number of candidate sets evaluated, and agreement with exhaustive enumeration. When exhaustive enumeration is feasible, it is used to distinguish scoring failure from approximate decoding failure. When exhaustive enumeration becomes infeasible, we report runtime and recovery trade-offs.

The central question is whether scalable decoding can preserve most of the recovery achieved by exhaustive search while reducing the combinatorial cost.

\subsection{Experiment 4: Representative Score-Gap and True-Set-Rank Diagnostics}
\label{app:oracle_scoregap_rank}

This appendix provides a representative seed-42 score-gap and true-set-rank diagnostic under oracle retrieval. 

\begin{table*}[t]
\centering
\small
\caption{Score-gap and true-set-rank validation under oracle retrieval on SERGIO DS3, seed 42. Targets are grouped by exact-recovery outcome. Exact recovery occurs when the true set receives the highest score, corresponding to positive score gaps and true-set rank 1. Failed exact recovery corresponds to negative score gaps and true-set ranks greater than 1.}
\label{tab:appendix-scoregap-by-outcome}
\begin{tabular}{clcccccc}
\toprule
\(R\) & Outcome & Targets & Mean Gap & Median Gap & Mean True Rank & Median True Rank & Mean Jaccard \\
\midrule
2 & Failed exact recovery & 41 & -4.480 & -2.769 & 214.195 & 12.0 & 0.179 \\
2 & Exact recovery & 16 & 1.197 & 1.181 & 1.000 & 1.0 & 1.000 \\
\midrule
3 & Failed exact recovery & 26 & -3.515 & -2.994 & 364.846 & 28.0 & 0.323 \\
3 & Exact recovery & 13 & 0.971 & 0.831 & 1.000 & 1.0 & 1.000 \\
\midrule
4 & Failed exact recovery & 15 & -7.565 & -6.699 & 5006.600 & 689.0 & 0.290 \\
4 & Exact recovery & 4 & 0.859 & 0.897 & 1.000 & 1.0 & 1.000 \\
\bottomrule
\end{tabular}
\end{table*}

It is intended to illustrate the set-level scoring failure mechanism, while the main oracle-retrieval and TRACE attribution results are reported across five seeds. In this setting, oracle retrieval forces the true regulator set to be present in the candidate pool. Therefore, negative score gaps and true-set ranks greater than 1 indicate that the true set was available but scored below an incorrect alternative. 

For each target gene \(t\), the score gap is defined as
\[
\mathrm{Gap}_t =
\mathrm{Score}(S_t,t)
-
\mathrm{Score}(S_t^{-},t),
\]
where \(S_t\) is the true regulator set and \(S_t^{-}\) is the highest-scoring
incorrect candidate set. A positive score gap indicates that the true regulator
set receives a higher score than every incorrect candidate set in the evaluated
pool. A negative score gap indicates that at least one incorrect set is scored
above the true set. We also compute the true-set rank, where rank 1 means that
the true regulator set is the highest-scoring set.

Table~\ref{tab:appendix-oracle-scoregap-rank} reports aggregate score-gap and
true-set-rank diagnostics under oracle retrieval.

\begin{table}[H]
\centering
\footnotesize
\setlength{\tabcolsep}{2.4pt}
\renewcommand{\arraystretch}{1.12}
\caption{Oracle-retrieval score-gap and true-set-rank diagnostics for Residual HOS2 on SERGIO DS3, seed 42.}
\label{tab:appendix-oracle-scoregap-rank}
\begin{tabular}{@{}cccccccc@{}}
\toprule
\(R\) 
& \(M\) 
& Targets 
& \shortstack{Oracle\\Exact} 
& \shortstack{Mean\\Gap} 
& \shortstack{Median\\Gap} 
& \shortstack{Mean\\Rank} 
& \shortstack{Median\\Rank} \\
\midrule
2 & 80 & 57 & 0.281 & -2.887 & -1.309 & 154.351 & 5 \\
3 & 80 & 39 & 0.333 & -2.020 & -1.466 & 243.564 & 3 \\
4 & 55 & 19 & 0.211 & -5.791 & -5.721 & 3952.789 & 486 \\
\bottomrule
\end{tabular}
\end{table}

\begin{figure}[t]
\centering
\includegraphics[width=0.45\linewidth]{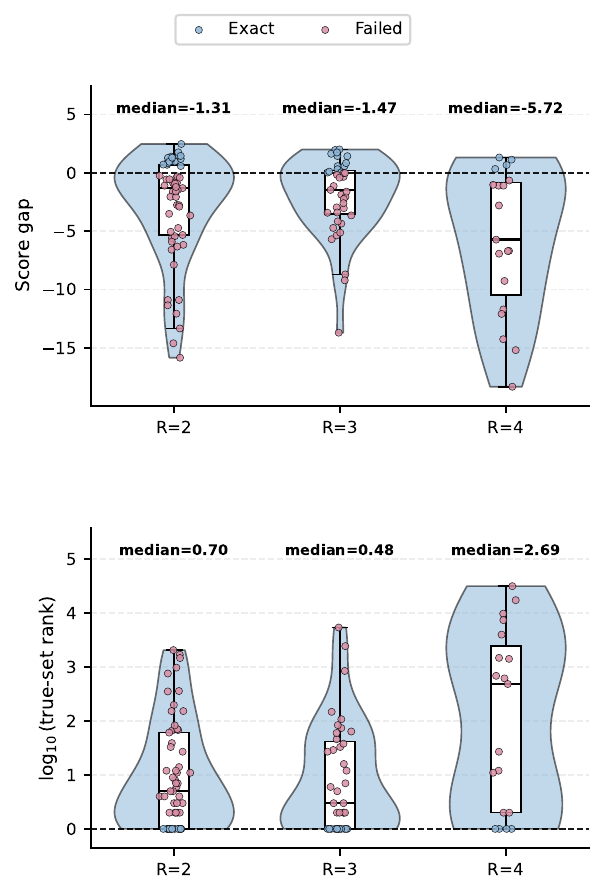}
\caption{
Per-target score-gap and true-set-rank distributions under oracle retrieval on
SERGIO DS3, seed 42. The top panel shows score-gap distributions across targets
for each regulator-set size. The dashed horizontal line marks zero; values below
zero indicate that the true regulator set is present in the candidate pool but
is scored below at least one incorrect set. The bottom panel shows
\(\log_{10}\) true-set rank, where zero corresponds to rank 1. Points are
colored by exact-recovery outcome. Exact indicates that the true regulator set
is ranked first; failed indicates that at least one incorrect candidate set is
ranked above the true set.
}
\label{fig:oracle-scoregap-rank-violin}
\end{figure}

Table~\ref{tab:appendix-oracle-scoregap-rank} shows that the true regulator set
is often misranked even when oracle retrieval guarantees that it is present in
the candidate pool. The mean and median score gaps are negative for all
regulator-set sizes, indicating that the strongest incorrect set often receives
a higher HOS2 score than the true set. This effect is especially severe for
\(R=4\), where the median true-set rank is 486. These diagnostics support the
TRACE interpretation that remaining oracle-retrieval failures are primarily
scoring failures rather than retrieval failures.
Figure~\ref{fig:oracle-scoregap-rank-violin} provides a distribution-level view
of the score-gap and true-set-rank diagnostics. While
Table~\ref{tab:appendix-oracle-scoregap-rank} reports aggregate means and
medians, the violin plot shows how these failures vary across individual target
genes. This is important because oracle retrieval guarantees that the true
regulator set is already present in the candidate pool. Therefore, negative
score gaps and true-set ranks greater than one cannot be explained by missing
candidates; they indicate that the scorer ranks an incorrect regulator set
above the true set.

In the figure, ``Exact'' denotes targets for which the true regulator set is
ranked first and therefore exactly recovered. ``Failed'' denotes targets for
which the true regulator set is present but not selected as the top-scoring
prediction. The separation between exact and failed targets validates the
diagnostic: exact cases concentrate at positive score gaps and rank 1, whereas
failed cases concentrate below the zero-gap threshold and above rank 1.

Table~\ref{tab:decoder-scalability-5seed} shows that PairS2 proposal followed by HOS2 reranking substantially reduces decoding cost while preserving exact recovery across five random seeds. For $R=2$, the proposal decoder evaluates only 190 candidate sets per target instead of 3,160, reducing the search space by 93.99\% with only a negligible decrease in exact recovery. For $R=3$, the proposal decoder evaluates 2,300 sets instead of 82,160 and improves exact recovery from 0.3179 to 0.3333 while reducing the search space by 97.20\%. For $R=4$, exact recovery is unchanged at 0.2211 while the number of scored sets decreases from 341,055 to 12,650, corresponding to a 96.29\% reduction.

These results indicate that proposal-based reranking is an effective decoding intervention within BRIDGE. Unlike hard-negative fine-tuning, which modifies the learned HOS2 scorer and can overcorrect, PairS2 proposal followed by HOS2 reranking keeps the HOS2 scorer fixed and reduces only the candidate set space considered at inference time. This makes proposal reranking a safe scalability intervention: it substantially reduces search cost while preserving the recovery behavior of exhaustive HOS2 decoding.

\begin{figure}[H]
\centering
\includegraphics[width=0.62\linewidth]{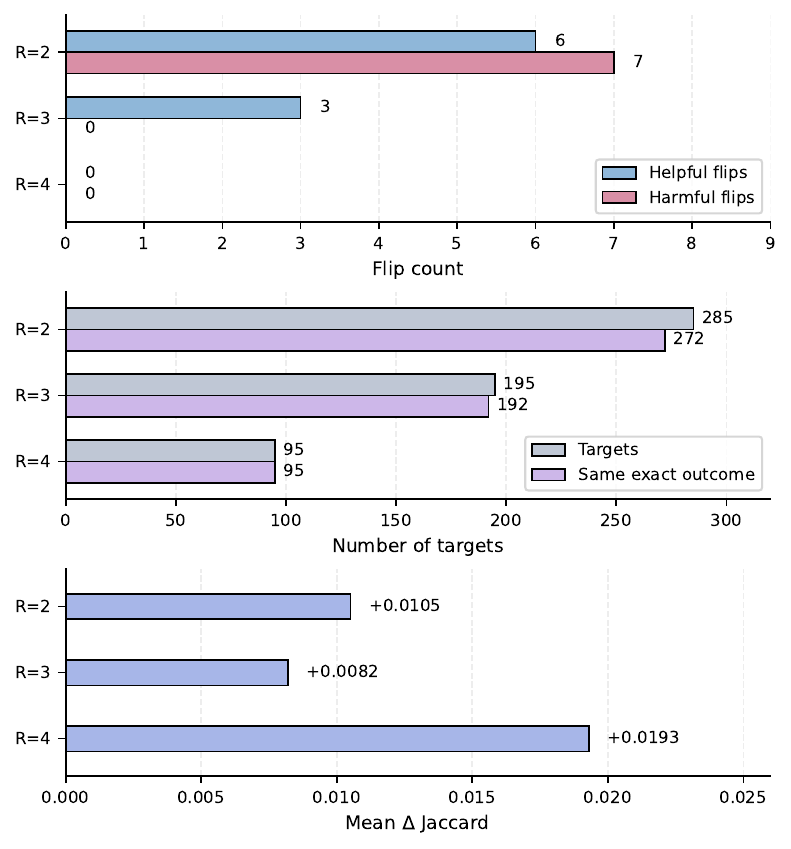}
\caption{
Five-seed exact-outcome flip analysis comparing exhaustive HOS2 decoding with
PairS2 proposal followed by HOS2 reranking. The top panel shows helpful and
harmful flips, the middle panel compares total targets with unchanged exact
outcomes, and the bottom panel shows mean change in Jaccard similarity.
Helpful flips indicate failures corrected by proposal reranking; harmful flips
indicate correct exhaustive predictions lost after proposal reranking. 
}
\label{fig:decoder-flip-analysis}
\end{figure}

Figure~\ref{fig:decoder-flip-analysis} provides a target-level exact-outcome analysis comparing proposal reranking against exhaustive HOS2 decoding. For $R=2$, proposal reranking yields six helpful flips and seven harmful flips across 285 evaluated targets, producing a nearly neutral net change in exact recovery. All seven harmful flips occur when the true regulator set is no longer fully contained in the reduced proposal pool, indicating that these failures are caused by proposal-stage pruning rather than by HOS2 reranking. For $R=3$, proposal reranking yields three helpful flips and no harmful flips, explaining the slight improvement in exact recovery. For $R=4$, no exact-outcome flips occur, showing that proposal reranking preserves exhaustive HOS2 exact behavior while reducing the search space by 96.29\%.

Together, the scalability and flip analyses show that proposal-based reranking dramatically reduces the number of HOS2-scored candidate sets, preserves exact recovery for nearly all targets, and produces rare harmful changes only when the reduced proposal pool removes a true regulator. BRIDGE therefore accepts proposal-based reranking as a safe decoding intervention while treating proposal coverage as the remaining failure mode to monitor.

A seed-42 score-based decoding-loss audit is included in
Appendix~\ref{app:score_based_decoding_loss}.

\subsubsection{Validation by Exact-Recovery Outcome}

Table~\ref{tab:appendix-scoregap-by-outcome} validates the score-gap diagnostic
by grouping oracle-retrieval targets according to exact-recovery outcome.

Across all regulator-set sizes, exact recovery corresponds to positive score
gaps and true-set rank 1, meaning that the true regulator set is the
highest-scoring set in the oracle candidate pool. In contrast, failed exact
recovery corresponds to negative score gaps and true-set ranks greater than 1,
indicating that at least one incorrect candidate set receives a higher HOS2
score than the true set. The effect becomes especially severe for \(R=4\), where
failed targets have a median true-set rank of 689. These results support the use
of score gap and true-set rank as direct diagnostics of set-level scoring
failure.

\subsection{Experiment 6: TRACE Failure Attribution}
\label{app:trace-failure-attribution}

To isolate the dominant source of recovery failure, we decompose performance into retrieval, scoring, and decoding components.

For each target, we compute coverage, conditional exact recovery, and final exact recovery.

We estimate retrieval loss as:

\[
\mathrm{RetrievalLoss}
=
1-\mathrm{Coverage}.
\]

\[
\begin{aligned}
\mathrm{ScoringLoss}
&=
\mathrm{Coverage}\times(1-\mathrm{ConditionalExact}) \\
&=
\mathrm{Coverage}-\mathrm{UnconditionalExact}.
\end{aligned}
\]
To quantify the severity of scoring failure, we also report the score gap and true-set rank when exhaustive scoring is feasible.

When exhaustive decoding is used, decoding loss is zero by construction because all candidate subsets in the capped pool are evaluated. For approximate decoding, we define decoding loss relative to exhaustive enumeration:

\[
\mathrm{DecodingLoss}
=
\Pr\left(
\mathrm{Score}(\widehat{S}^{\mathrm{approx}}_t,t)
<
\mathrm{Score}(\widehat{S}^{\mathrm{exact}}_t,t)
\right).
\]

The central question is which stage contributes most to recovery failure.
\begin{table}[H]
\centering
\footnotesize
\setlength{\tabcolsep}{2.4pt}
\renewcommand{\arraystretch}{1.12}
\caption{TRACE failure attribution for Residual HOS2 on SERGIO DS3. Results are reported as mean \(\pm\) standard deviation across seeds \(\{42,43,44,45,46\}\).}
\label{tab:trace-failure-attribution-five-seed}
\begin{tabular}{@{}ccccccc@{}}
\toprule
\(R\) 
& Cov. 
& Exact 
& \shortstack{Ret.\\Loss} 
& \shortstack{Score\\Loss} 
& \shortstack{Dec.\\Loss} 
& Bottleneck \\
\midrule
2 & \shortstack{0.874\\\(\pm 0.040\)} & \shortstack{0.298\\\(\pm 0.066\)} & \shortstack{0.126\\\(\pm 0.040\)} & \shortstack{0.575\\\(\pm 0.079\)} & \shortstack{0.000\\\(\pm 0.000\)} & Scoring \\
\midrule
3 & \shortstack{0.938\\\(\pm 0.014\)} & \shortstack{0.318\\\(\pm 0.050\)} & \shortstack{0.062\\\(\pm 0.014\)} & \shortstack{0.621\\\(\pm 0.061\)} & \shortstack{0.000\\\(\pm 0.000\)} & Scoring \\
\midrule
4 & \shortstack{0.863\\\(\pm 0.071\)} & \shortstack{0.221\\\(\pm 0.058\)} & \shortstack{0.137\\\(\pm 0.071\)} & \shortstack{0.642\\\(\pm 0.086\)} & \shortstack{0.000\\\(\pm 0.000\)} & Scoring \\\bottomrule
\end{tabular}
\end{table}
Table~\ref{tab:trace-failure-attribution-five-seed} shows that scoring is the dominant bottleneck across all regulator-set sizes. For \(R=2\), retrieval loss is \(0.126\pm0.040\), while scoring loss is \(0.575\pm0.079\). For \(R=3\), retrieval loss is \(0.062\pm0.014\), while scoring loss is \(0.621\pm0.061\). For \(R=4\), retrieval loss is \(0.137\pm0.071\), while scoring loss is \(0.642\pm0.086\). Since exhaustive decoding is used, decoding loss is zero by construction. Thus, TRACE attributes the dominant remaining failure mode to set-level scoring rather than retrieval or approximate decoding.
\subsection{Experiment 7: BRIDGE Bottleneck-Aware Recovery Study}
\label{app:bridge-intervention-policy}
This experiment evaluates the full BRIDGE framework as a
bottleneck-aware intervention-selection procedure. Rather than applying one
global modification, BRIDGE uses TRACE diagnoses and held-out validation to
determine whether a targeted intervention should be accepted or rejected.

We evaluate two intervention families. For scoring repair, we test
hard-negative-enhanced HOS2 and a seed-42 model-mined confuser fine-tuning
pilot. For decoding scalability, we test PairS2-guided proposal followed by
Residual HOS2 reranking. In the decoding intervention, PairS2 ranks the
regulators in the capped Stage-1 pool and retains the top-$L$ candidates, where
$L=\{20,25,25\}$ for $R=\{2,3,4\}$, respectively. Residual HOS2 then
exhaustively evaluates all size-$R$ subsets within this reduced proposal pool.
Thus, the decoder is exhaustive within the reduced pool but approximate
relative to exhaustive HOS2 decoding over the original candidate pool.

BRIDGE accepts an intervention only when it improves or safely preserves
held-out recovery while addressing the diagnosed bottleneck or computational
constraint. We report Coverage, Conditional Exact Match, Final Exact Match,
Recall, Jaccard Similarity, Score Gap, Runtime, and Number of Candidate Sets
Evaluated. The central question is whether TRACE-guided intervention selection
can improve recovery or scalability without overcorrecting the model.

The intervention studies show that BRIDGE should not be interpreted as a single global retraining rule. Instead, BRIDGE uses TRACE diagnostics and intervention validation to decide whether an intervention should be applied or rejected. Since scoring is the dominant bottleneck for all regulator-set sizes, we evaluated increasingly strong scoring interventions. However, the effect of these interventions was size-dependent.

Table~\ref{tab:bridge-policy} summarizes the resulting BRIDGE intervention
policy. For scoring interventions, BRIDGE selects hard-negative-enhanced
training only for \(R=2\), where it improves held-out exact recovery across
seeds. A seed-42 model-mined fine-tuning pilot also improves \(R=2\) exact
recovery from 0.263 to 0.316. However, for \(R=3\) and \(R=4\), stronger
hard-negative pressure degrades recovery despite the presence of a scoring
bottleneck. The seed-42 model-mined fine-tuning pilot further decreases exact
recovery from 0.333 to 0.231 for \(R=3\) and from 0.211 to 0.105 for \(R=4\).
BRIDGE therefore rejects these scoring interventions for larger regulator sets
and retains the baseline HOS2 scorer.

In contrast, the decoding intervention is accepted as a scalability
intervention across all regulator-set sizes. PairS2-guided proposal followed by
Residual HOS2 reranking largely preserves exact-recovery behavior while
reducing the number of HOS2-scored candidate sets by 93.99\%, 97.20\%, and
96.29\% for $R=2$, $R=3$, and $R=4$, respectively. Thus, BRIDGE functions not only as a recovery framework but also as an intervention-selection safeguard: it accepts targeted interventions that improve or safely preserve recovery and rejects interventions that overcorrect the model.

This policy-level interpretation is important because the same TRACE diagnosis
can lead to different accepted actions depending on validation behavior. In all
three regulator-set sizes, TRACE identifies scoring as the dominant bottleneck,
but the validation experiments show that scoring repair is not uniformly safe.
For \(R=2\), the candidate set space is smaller and near-miss negatives are
easier to separate, so hard-negative-enhanced scoring can improve exact
recovery. For \(R=3\) and \(R=4\), the set space is larger and many incorrect
sets share multiple regulators with the true set, making aggressive
confuser-based repair more likely to overfit to local ranking errors. The
decoding intervention behaves differently: PairS2 proposal followed by HOS2
reranking changes the search procedure without changing the learned HOS2
scorer, so it reduces computational burden while preserving the underlying
scoring behavior. This distinction explains why BRIDGE accepts proposal-based
reranking across all regulator-set sizes but accepts scoring repair only in the
validated \(R=2\) setting.
\begin{table*}[t]
\centering
\small
\setlength{\tabcolsep}{4pt}
\renewcommand{\arraystretch}{1.25}
\caption{TRACE-guided BRIDGE intervention policy. BRIDGE selects interventions
only when they improve or safely preserve held-out recovery while addressing a
diagnosed bottleneck or computational constraint.}
\label{tab:bridge-policy}
\begin{tabularx}{\textwidth}{@{}>{\raggedright\arraybackslash}p{0.14\textwidth}
>{\raggedright\arraybackslash}p{0.17\textwidth}
>{\raggedright\arraybackslash}X
>{\raggedright\arraybackslash}p{0.22\textwidth}@{}}
\toprule
Setting & TRACE Diagnosis & Validation Outcome & BRIDGE Action \\
\midrule
$R=2$ scoring
& Scoring bottleneck
& Hard-negative-enhanced HOS2 increases the three-seed mean exact recovery
from 0.2865 to 0.3216, but the effect is inconsistent across seeds and is
not statistically significant. In the seed-42 pilot, model-mined fine-tuning
increases exact recovery from 0.263 to 0.316.
& Retain baseline HOS2 as the default; treat scoring repair as provisional
pending additional validation. \\
\midrule
$R=3$ scoring
& Scoring bottleneck
& Hard-negative-enhanced training decreases exact recovery relative to baseline.
The seed-42 model-mined fine-tuning pilot further decreases exact recovery from
0.333 to 0.231.
& Reject scoring intervention; retain baseline HOS2 scorer. \\
\midrule
$R=4$ scoring
& Scoring bottleneck
& Hard-negative-enhanced training decreases exact recovery relative to baseline.
The seed-42 model-mined fine-tuning pilot further decreases exact recovery from
0.211 to 0.105, suggesting overcorrection for larger regulator sets.
& Reject scoring intervention; retain baseline HOS2 scorer. \\
\midrule
All $R$ decoding
& Decoding-cost constraint
& PairS2-guided proposal followed by Residual HOS2 reranking largely preserves
exact-recovery behavior while reducing the number of HOS2-scored candidate
sets by 93.99\%, 97.20\%, and 96.29\% for $R=2,3,4$.
& Accept proposal-based reranking as a safe scalability intervention. \\
\bottomrule
\end{tabularx}
\end{table*}
\FloatBarrier

\begin{figure}[t]
    \centering
    \includegraphics[
        width=0.62\linewidth
    ]{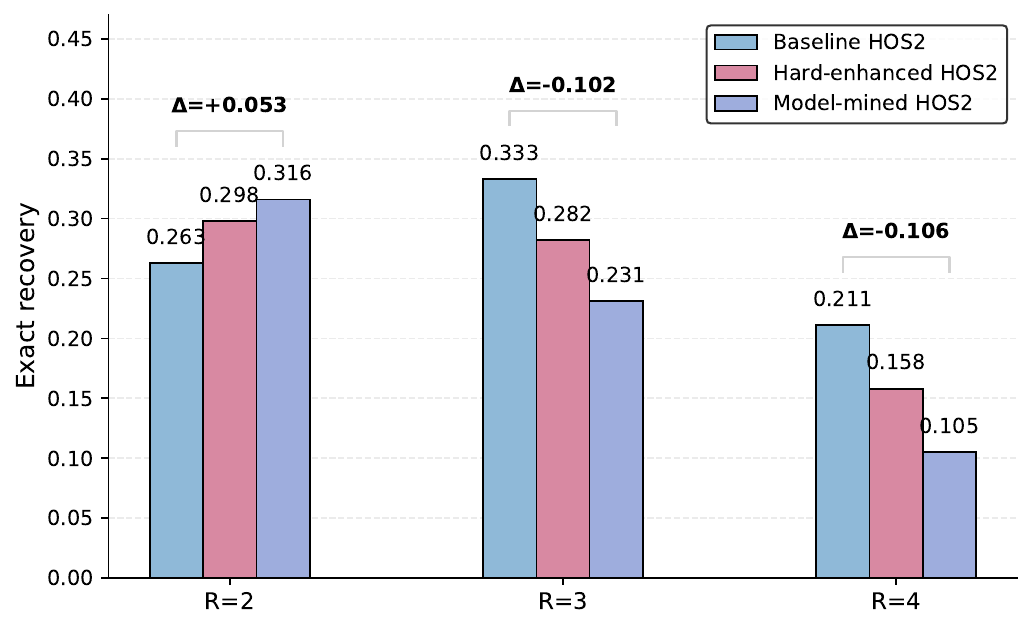}
    \caption{
    Seed-42 pilot scoring-intervention results for exact regulator-set
    recovery across regulator-set sizes. Bars show exact recovery for the original HOS2 scorer,
    hard-negative-enhanced HOS2, and model-mined HOS2. The bracketed
    \(\Delta\) values report the change from baseline HOS2 to model-mined HOS2.
    Model-mined fine-tuning improves exact recovery for \(R=2\) but reduces
    exact recovery for \(R=3\) and \(R=4\), indicating that scoring refinement
    alone does not uniformly resolve the higher-order recovery bottleneck.
    }
    \label{fig:scoring_intervention_pilot}
\end{figure}

\FloatBarrier

Figure~\ref{fig:scoring_intervention_pilot} summarizes the seed-42
scoring-intervention pilot. In this single-seed analysis, the
hard-negative-enhanced scorer does not consistently increase exact
recovery relative to baseline HOS2, while model-mined confuser
fine-tuning has a mixed effect: it raises exact recovery for \(R=2\)
but lowers it for \(R=3\) and \(R=4\). This suggests that directly
fine-tuning on model-mined confusers can overcorrect the scorer for
larger regulator sets rather than provide a uniformly beneficial
scoring repair. These pilot results should be interpreted separately
from the three-seed hard-negative analysis in
Appendix~\ref{app:hard-negative-intervention}.

\subsection{Experiment 8: TRACE-lite Comparison with External Edge-Ranking Baselines}

We evaluate external edge-ranking baselines using TRACE-lite, an operational version of TRACE for methods that do not expose explicit retrieval, set-scoring, and decoding modules. Correlation, mutual information, and GRNBoost2 produce regulator--target rankings rather than cooperative regulator sets. We therefore convert each method's target-specific ranking into a top-\(M\) candidate pool and a top-\(R\) predicted regulator set.

For each target gene \(t\), targets are grouped by the cardinality of their ground-truth regulator set, \(R=|S_t|\). The top-\(R\) ranked regulators are treated as the predicted set, and the top-\(M\) ranked regulators are treated as the retrieval pool. We use \(M=80\) for \(R=2,3\) and \(M=55\) for \(R=4\), matching the main BRIDGE/HOS2 evaluation protocol. No score or importance cutoff is applied; TRACE-lite is rank-based.

Correlation and mutual information provide simple pairwise association
baselines, while GRNBoost2 and GENIE3 provide tree-based GRN
edge-ranking baselines. On SERGIO DS3, GRNBoost2 produced 1,402,065 regulator--target edges
after self-edge removal, and all baseline rankings were evaluated under the
same TRACE-lite protocol.
The central question is whether external edge-ranking methods can recover complete cooperative regulator sets, or whether they primarily recover partial regulator--target signal.
\subsection{Planned Biological Plausibility Validation}
\label{app:planned-biological-validation}

Because complete ground-truth cooperative regulator sets are generally
unavailable in real single-cell datasets, this study uses controlled benchmarks
for exact regulator-set recovery and real-data experiments for TRACE-lite and
Stage-1 retrieval validation. A natural future validation is to apply BRIDGE to
real single-cell expression data and test whether recovered regulator--target
relationships are enriched for external biological evidence, such as curated
TF--target databases, pathway annotations, or perturbation-derived regulatory
effects.

This section is therefore a planned validation protocol rather than a completed
result. Such an analysis would not treat curated databases as complete ground
truth. Instead, it would ask whether BRIDGE-recovered regulator sets are more
biologically plausible than sets recovered from edge-ranking baselines. We leave
this biological-plausibility validation to future work.
\subsection{Additional Information-Theoretic Extensions}
\label{app:information-theoretic-extensions}

The main paper discusses foundation-model and information-theoretic extensions
at a high level. A fuller treatment could estimate mutual information between
learned representations and regulator-set identities, regulator-set synergy with
target expression, or spectral entropy of learned embeddings. Because these
analyses depend strongly on estimator choices such as discretization,
k-nearest-neighbor, neural, or contrastive mutual-information estimators, we
leave them to future work and focus here on directly measurable BRIDGE
diagnostics.

\end{document}